\documentclass[11pt]{article}
\usepackage[final]{acl}

\usepackage{times}
\usepackage{latexsym}
\usepackage[T1]{fontenc}
\usepackage[utf8]{inputenc}
\usepackage{microtype}
\usepackage{inconsolata}
\usepackage{graphicx}
\usepackage{tcolorbox}
\tcbuselibrary{skins, breakable}
\usepackage{booktabs}
\usepackage{multirow}
\usepackage{adjustbox}
\usepackage{amsmath, amsfonts}
\usepackage{xcolor}
\usepackage{enumitem}
\usepackage{seqsplit}
\usepackage{tabularx}  
\usepackage{colortbl}  
\usepackage{subcaption}
\usepackage{wrapfig}

\usepackage[dvipsnames,svgnames]{xcolor}
\definecolor{mimic}{HTML}{E09E00}   
\definecolor{parkinson}{HTML}{008AE0}
\title{PACE-RAG: Patient-Aware Contextual and \\
Evidence-Constrained RAG for Clinical Drug Recommendation}

\author{
\textbf{Chaeyoung Huh}$^{1,*}$ \quad
\textbf{Hyunmin Hwang}$^{1,*}$ \quad  \\
\textbf{Jung Hwan Shin}$^{2}$ \quad 
\textbf{Sungyang Jo}$^{3}$ \quad
\textbf{Jinse Park}$^{4,\dagger}$ \quad
\textbf{Jong Chul Ye}$^{1,\dagger}$ \\
$^{1}$Graduate School of AI, Korea Advanced Institute of Science and Technology\\
$^{2}$Department of Neurology, Seoul National University Hospital \\
$^{3}$Department of Neurology, Asan Medical Center, University of Ulsan College of Medicine \\
$^{4}$Department of Neurology, Haeundae Paik Hospital, Inje University \\
\texttt{lirisnoir@kaist.ac.kr} \quad 
\texttt{hyunmin\_hwang@kaist.ac.kr} \quad  
\texttt{neo2003@snu.ac.kr} \\ 
\texttt{sungyangjo@amc.seoul.kr} \quad
\texttt{loca99@paik.ac.kr} \quad
\texttt{jong.ye@kaist.ac.kr}  \\
}

\begin{document}
\maketitle
\renewcommand{\thefootnote}{\fnsymbol{footnote}}
\footnotetext[1]{Equal contribution.}
\footnotetext[2]{Corresponding author.}

\begin{abstract}
Drug recommendation requires a deep understanding of individual patient context, especially for complex conditions like Parkinson’s disease. 
While LLMs possess broad medical knowledge, they fail to capture the subtle nuances of actual prescribing patterns. 
Existing RAG methods also struggle with these complexities because guideline-based retrieval remains too generic and similar-patient retrieval often replicates majority patterns without 
accounting for the unique clinical nuances of individual patients.
To bridge this gap, we propose \textbf{PACE-RAG} (\textbf{P}atient-\textbf{A}ware \textbf{C}ontextual and \textbf{E}vidence-Constrained 
\textbf{RAG}).
Rather than directly copying frequent medications from retrieved patients, PACE-RAG personalizes recommendations by first extracting patient-specific clinical features, retrieving cases around these features, and then refining the final prescription using the patient's current symptoms, active medication history, and focus-specific prescribing tendencies.
By analyzing treatment patterns tailored to specific clinical features,
PACE-RAG generates patient-specific medication recommendations along with an explainable clinical summary.
PACE-RAG achieved the strongest performance among the evaluated inference-only LLM-based methods, reaching F1 scores of 80.84\% and 47.22\% on the Parkinson's disease and MIMIC-IV cohorts, respectively.
Our code is available at: \url{https://github.com/ChaeYoungHuh/PACE-RAG}.

\end{abstract}

\section{Introduction}
\begin{figure}[t!]
    \centering
    \includegraphics[
        width=0.98\linewidth,
        trim=0cm 4cm 3.5cm 0cm,
        clip
    ]{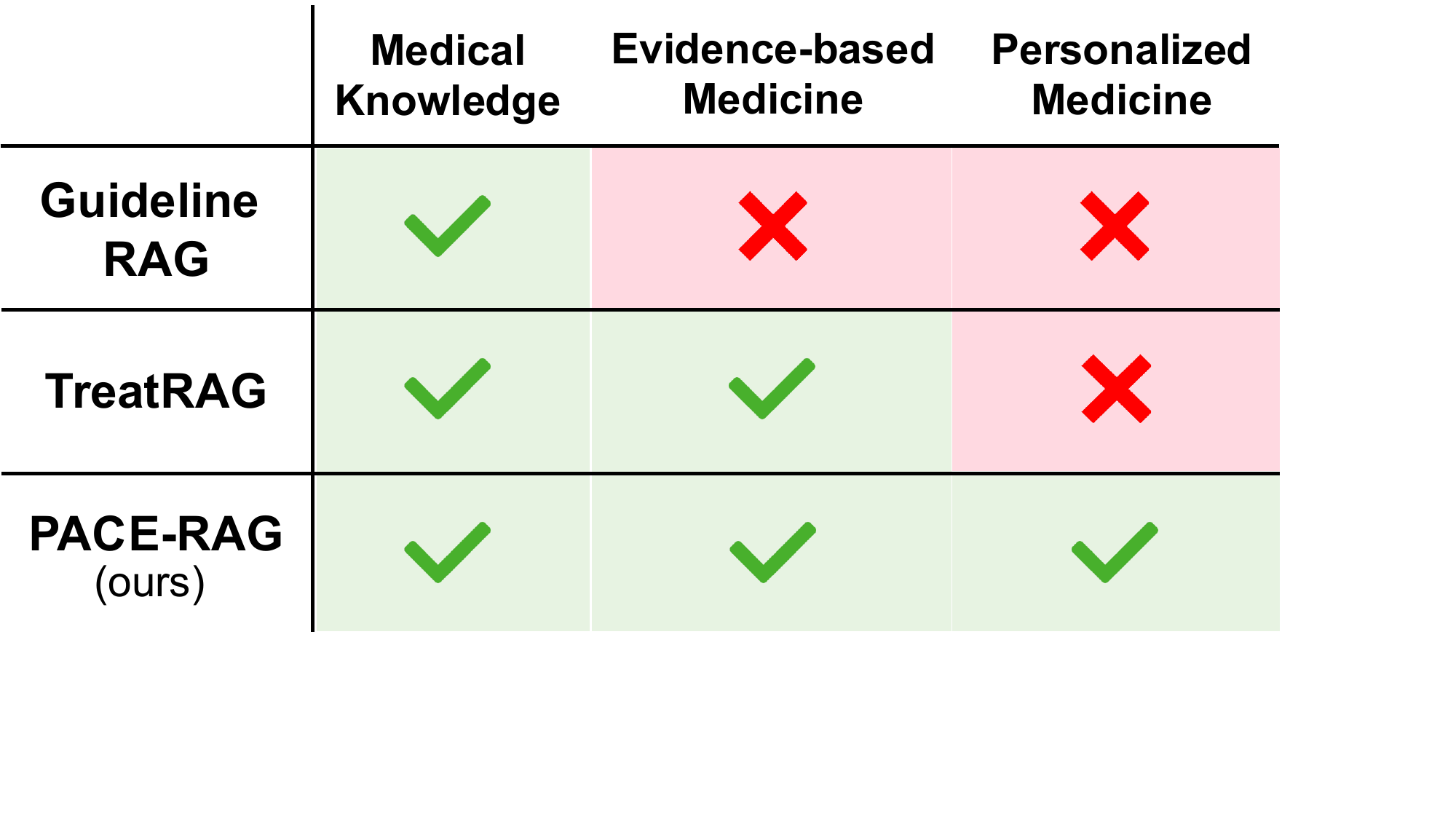}
\caption{
\textbf{Comparison of medical RAG approaches.}
Guideline RAG provides general medical knowledge, whereas TreatRAG retrieves similar patient cases but remains limited in personalization.
\textbf{PACE-RAG} achieves personalized medicine by integrating a verification process with patient-similarity retrieval.
}
    \label{fig:concept_figure}
    \vspace{-2em}
\end{figure}

\begin{figure*}[t]
    \centering
    \includegraphics[
        width=0.95\textwidth,
        trim=0cm 6.5cm 3cm 0cm,
        clip
    ]{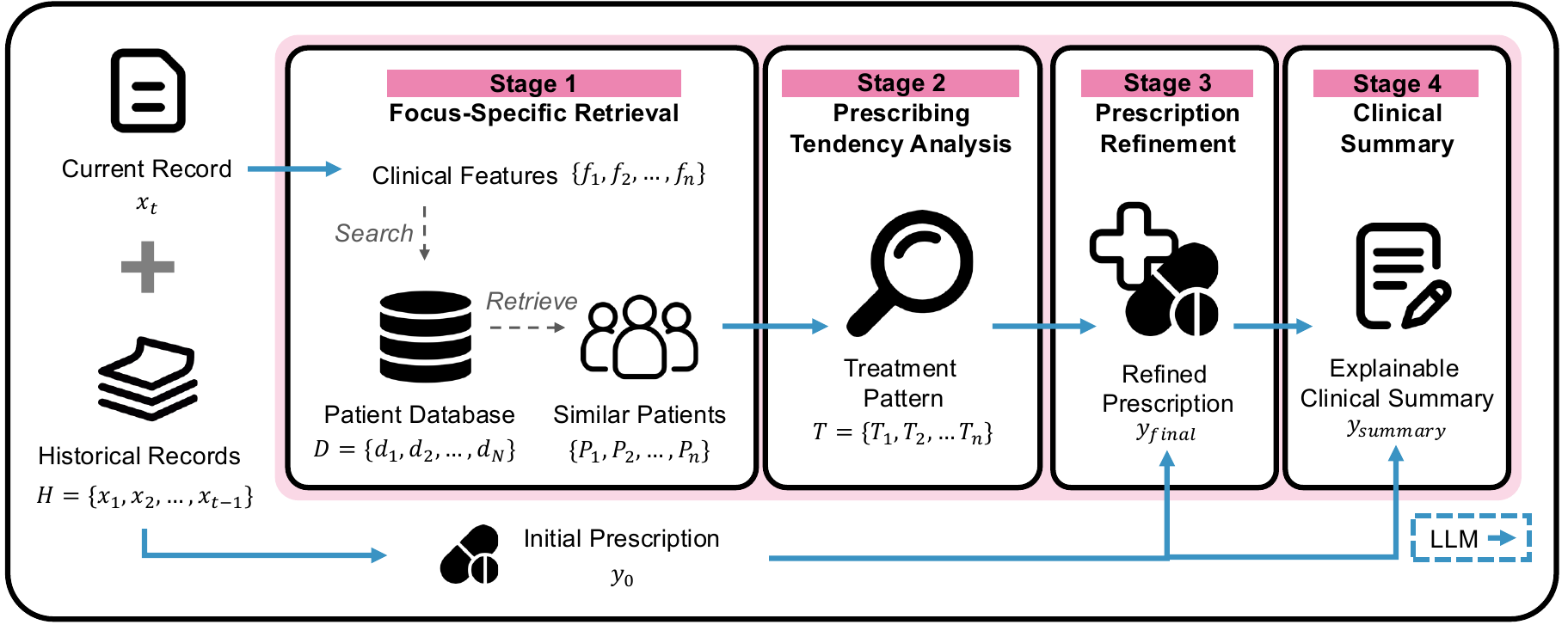}
    \caption{\textbf{Overview of the PACE-RAG framework architecture.} 
    The pipeline consists of four stages: (1) Focus-Specific Retrieval to identify similar patients from $x_t$; (2) Prescribing Tendency Analysis to extract medication patterns from retrieved cases; (3) Prescribing Refinement to verify the initial prescription against identified tendencies; and (4) Clinical Summary to synthesize the final treatment rationale.
    The LLM performs each generation step, from focus extraction to final summary generation.
    }
    \label{fig:framework}
\vspace{-1.8em}
\end{figure*}

Drug recommendation represents a complex clinical challenge that necessitates a highly personalized strategy tailored to the unique clinical presentation of each patient~\cite{nih_pd_review_2025}. 
Although Large Language Models (LLMs) possess extensive medical knowledge and achieve expert-level performance on standardized benchmarks~\cite{singhal2023large,gpt4_med}, they frequently struggle with the nuanced, context-aware decision-making required in complex clinical practice~\cite{deik_2023_perils}.
While physicians derive clinical judgments by synthesizing a patient's unique clinical condition with their own accumulated clinical experience, 
LLMs struggle to capture this real-world experiential knowledge.
Therefore, it is essential to develop methodologies that effectively incorporate practical clinical insights.

To address this, 
Retrieval-Augmented Generation (RAG) frameworks utilize external knowledge retrieved from clinical guidelines or medical textbooks~\cite{wang_2025_medrag, wang2024apollo}. 
However, these methods can suffer from semantic fragmentation during document chunking and fail to capture the intuitive judgment characteristic of expert physicians.
As illustrated in Figure~\ref{fig:concept_figure}, {\em Guideline RAG}, which retrieves standardized clinical documents containing standardized disease management guidelines, generates overly generalized reasoning that overlooks heterogeneous patient trajectories~\cite{armstrong_2020_pd}.
To overcome this rigidity, recent research retrieved similar historical patient cases~\cite{rag_survey, treatrag_2025, chen2026medcopilotmedicalassistantpowered}. 
While retrieving similar patient treatment cases ({\em Similar Patient RAG}) such as TreatRAG~\cite{treatrag_2025}  offers more specific analysis than Guideline RAG, it remains susceptible to `majority bias,' 
in which the model simply replicates frequent prescription patterns rather than identifying case-specific clinical rationales.
The major limitation of this approach is that patient-level similarity alone does not guarantee personalized prescribing.

To address these limitations, we propose \textbf{PACE-RAG} (\textbf{P}atient-\textbf{A}ware \textbf{C}ontextual and \textbf{E}vidence-Constrained RAG), 
a novel framework designed to capture each patient's context and leverage evidence-based retrieval to model the implicit prescribing 
tendencies
of clinical experts. 
As shown in Fig.~\ref{fig:framework}, our framework employs a four-stage reasoning process to identify 
patient-specific medication recommendations
for each patient based on their specific clinical presentation.

First, we employ \textit{Focus-Specific Retrieval}, which retrieves similar patients using symptom-level clinical features rather than the full clinical note, thereby focusing retrieval on patient-specific clinical signals.
Second, we implement \textit{Prescribing Tendency Analysis}. The model extracts treatment patterns from the retrieved cohort, effectively modeling how physicians typically respond to specific symptom clusters.
Third, we introduce 
\textit{Evidence-Constrained Prescription Refinement}, 
which verifies the initial medication plan against the prescribing tendencies derived from similar patient cases.
Finally, the \textit{Explainable Clinical Summary} synthesizes the decision trajectory into an interpretable report to ensure clinical trust and utility.

We evaluate our framework using a real-world Parkinson’s Disease dataset. 
Our results exhibit superior efficacy compared to alternative retrieval-based methods and verification frameworks.
To evaluate transfer to a substantially different structured clinical benchmark,
we extended our evaluation to the MIMIC-IV benchmark~\cite{johnson_2023_mimic_iv}. These results suggest that our method handles the complexities of multi-symptom clinical records.

In summary, our contributions are as follows:
\begin{itemize}
    \vspace{-1mm}
    \item We propose \textbf{PACE-RAG}, which mitigates the over-generalization of guideline-based retrieval and the majority-bias of similar-patient retrieval by identifying symptom-specific rationales within individual patient contexts.
    \item We introduce a four-stage methodology that extracts focus-specific clinical signals to perform keyword-centric retrieval of similar cases and analyzes prescribing tendencies to generate context-aware medication plans.
    \item We further evaluate the framework on MIMIC-IV, a substantially different structured clinical benchmark, demonstrating its applicability beyond the original Parkinson's disease setting.
\end{itemize}

\section{Related Work}
\vspace{-1mm}
\subsection{LLMs in Clinical Decision Support}
LLMs have demonstrated remarkable capabilities in the medical domain, achieving expert-level performance on benchmarks such as the USMLE and MedQA~\cite{singhal2023large, gpt4_med}. 
Beyond these standardized benchmarks, recent research has expanded the application of LLMs to complex clinical workflows, including differential diagnosis and treatment planning~\cite{kanjee_2023_diagnosis, thirunavukarasu_2023_llm, chen_2023_neurology}.
However, despite their medical reasoning capabilities,
LLMs struggle with \textit{personalized} and \textit{context-dependent} care, particularly in neurodegenerative disorders like Parkinson’s disease. 
For instance, while LLMs can accurately provide general knowledge regarding the symptoms of Parkinson’s disease—specifically movement disorders—they often fail to reflect the granular, patient-specific context required for real-world clinical prescriptions~\cite{deik_2023_perils}.
Our work addresses this limitation with a context-aware system that integrates an individual patient’s treatment trajectory with prescription patterns from similar patient cases.

\subsection{Retrieval Framework in Clinical Settings}
To mitigate domain-specific knowledge gaps, RAG has been widely adopted in the medical domain~\cite{rag_survey, xiong_2024_benchmark, asai2024selfrag, wang_2025_medrag, zakka_2024_almanac, treatrag_2025}. 
Existing approaches primarily utilize either explicit medical knowledge (Guideline-based) or historical patient cases (Case-based).
Guideline-based approaches prioritize clinical standardization by retrieving information from medical textbooks or clinical practice guidelines~~\cite{wang_2025_medrag, zakka_2024_almanac}. 
While effective at providing verified medical facts, the reliance on static knowledge-based retrieval cannot fully capture the heterogeneous clinical profiles and diverse situational nuances of individual patients in real-world practice.
Evidence-based retrieval instead utilizes historical EHR data to provide more tailored support, but existing models often rely on global semantic similarity that overlooks localized symptom signals~\cite{treatrag_2025}. 
Our approach introduces a novel framework that extracts primary clinical symptoms to drive targeted, keyword-based retrieval, enabling retrieval around patient-specific clinical signals.

\subsection{Verification of Clinical Reasoning}

While LLMs utilize RAG to minimize hallucinations, they are prone to an over-reliance on retrieved contexts. 
Consequently, a verification stage is essential to validate the model’s response and discern only the most meaningful insights from the retrieved content.
To address this, existing approaches prioritize internal consistency or introspection to refine model outputs~\cite{manakul_2023_selfcheckgpt, gandhi_2024_sos, li_2025_reflectevo}. 
Within the clinical domain, AMIE~\cite{tu2025amie} and MedAgents~\cite{tang2023medagents} have been further adapted to emulate the reasoning processes of medical experts. 
For instance, MedReflect~\cite{huang_2026_medreflect} minimizes errors by generating and training on question-answer pairs designed to validate reflective reasoning paths, thereby refining the model's internal logic.
However, these existing verification frameworks struggle to incorporate the unique clinical contexts of individual patients.
To address this, we introduce an 
Evidence-Constrained Prescription Refinement
stage which evaluates the initial prescription based on the prescribing tendencies derived from similar patient cases, 
preventing over-dependence on raw retrieved content by validating the treatment plan against essential clinical information.

\section{Methodology: PACE-RAG}
\label{sec:Methodology}
\vspace{-1mm}
We consider an outpatient prescription scenario, where physicians make medication decisions based on the patient's current symptoms and recent visit history.
At each visit, the decision is to determine whether to maintain existing medications, add new drugs for newly emerged or worsening symptoms, or discontinue inappropriate medications.
We denote the current clinical note as $x_t$ and define the patient's recent historical visit records as 
$H_t = \{x_{t-m}, \dots, x_{t-1}\}$, where $m \leq 3$ indicates that we use at most the three most recent visits before the current visit.
Given $H_t$ and $x_t$, our goal is to generate a prescription set $Y_t = \{y_1, \dots, y_k\}$, where each $y_i$ is an individual medication and $k \geq 1$.

\paragraph{Initial prescription generation.} 
Before the execution of the multi-stage reasoning pipeline, the LLM, denoted as $\phi$, produces an initial candidate prescription $y_0$ (prompt provided in Appendix~\ref{app:prompts_initial_draft}). 
This prescription is generated directly from the patient’s longitudinal history $H$ and the current clinical presentation $x_t$, as follows:
\begin{equation}
    y_0 = \phi(H, x_t)
\end{equation}
This initial output acts as an initial prior knowledge.
To refine this initial prior into a clinically precise prescription, the clinical plan is generated through a four-stage reasoning pipeline: Focus-Specific Retrieval ($\S$\ref{method:stage1}), Prescribing Tendency Analysis ($\S$\ref{method:stage2}), 
Prescription Refinement  ($\S$\ref{method:stage3}), 
and Clinical Summary ($\S$\ref{method:stage4}). Figure \ref{fig:framework} illustrates the overall workflow.

\subsection{Stage 1: Focus-Specific Retrieval}
\label{method:stage1}

\noindent\textbf{Focus query extractor.} 
To retrieve patients with similar symptoms, the Focus Query Extractor isolates the most salient features from the current clinical note $x_t$ by identifying $n$ distinct clinical queries. 
This process is defined as:
\vspace{-0.5em}
\begin{equation}
    \{f_1, f_2, ..., f_n\}=\phi(x_{t})
\vspace{-0.5em}
\end{equation}
where each $f_i$ represents a localized clinical feature within $x_t$.
To prevent hallucinations, each $f_i$ is strictly constrained to literal substrings found within $x_t$, ensuring the extractor captures only significant clinical indicators while filtering out overly broad or non-specific terms
(details in Appendix~\ref{app:data_extract_focus} and \ref{app:prompts_focus_query_extractor}).

\paragraph{Retrieving similar patients.}
We perform a retrieval across a historical patient database $D = \{d_1, d_2, \dots, d_N\}$.
For each clinical feature $f_i$,
we retrieve the top-$k$ most relevant patient cases among those that exceed a similarity threshold $\tau$.
We further examine the sensitivity to $\tau$ and the retrieval count $k$ in Appendix~\ref{app:hyperparamter_sensitivity}.
\vspace{-0.5em}
\begin{equation}
    P_i = S(f_i, D)
\vspace{-0.5em}
\end{equation}
where  $S(\cdot)$ denotes a selection function that identifies the nearest neighbors within the latent embedding space (details in Appendix~\ref{app:rag_detail}).
This produces one retrieved patient cluster $P_i$ for each clinical feature $f_i$.
This yields $n$ focus-specific patient clusters,
$P = \{P_1, P_2, \dots, P_n\}$.
This makes retrieval patient-specific because PACE-RAG retrieves evidence around the clinical features that are most relevant to the current patient's medication decision, rather than globally similar records based on the entire note.

\subsection{Stage 2: Prescribing Tendency Analysis}
\label{method:stage2}

Instead of analyzing the entire prescription history of retrieved patients, 
this stage conducts a targeted prescribing tendency analysis to isolate the direct impact of the 
clinical feature $f_i$.
This approach prevents over-generalized recommendations of high-frequency drugs that lack specific clinical evidence for the 
clinical feature.
For each clinical feature $f_i$,
the model analyzes the corresponding retrieved cluster $P_i$ to 
extract $T_i$, a set of medications specific to the 
feature
along with their underlying rationales.
To ensure high clinical fidelity and prevent hallucinations, the tendency analyzer $\phi$ enforces three rigorous criteria:
\begin{itemize}
\item 
    \textbf{Isolation of focus-specific drugs:} 
    The model extracts only medications directly relevant to $f_i$, 
    filtering out general or routine drugs to isolate 
    feature-specific interventions.
    \item \textbf{Evidence-based rationale:} 
    Each added medication must be supported by direct evidence relating to 
    the clinical feature $f_i$.
    \item \textbf{Null on uncertainty:} 
    In the absence of explicit evidence attributing a drug to $f_i$, the output $T_i$ returns to an empty set ($\emptyset$).
\end{itemize}

The total tendency set for the patient is defined as $T = \{T_1, T_2, \dots, T_n\}$, where each $T_i$ is derived 
from the corresponding clinical feature $f_i$ and retrieved patient cluster $P_i$
as follows (prompt in Appendix~\ref{app:prompts_tendency_analysis}):
\vspace{-0.5em}
\begin{equation}
T_{i} = \phi(f_i,P_i)
\vspace{-0.5em}
\end{equation}
By centering the analysis exclusively on 
feature-specific interventions,
this step identifies evidence-based treatment adjustments. 
These feature-specific tendency signals subsequently serve as evidence for refining whether to maintain, add, or remove medications in the next stage.
By extracting only feature-relevant prescribing tendencies, PACE-RAG converts similar-patient retrieval into patient-specific medication evidence.

\subsection{Stage 3: 
Evidence-Constrained Prescription Refinement}
\label{method:stage3}
If the initial draft ($y_0$) is directly incorporated with the retrieved context without a dedicated verification step, 
the model becomes overly dependent on the raw retrieved data. This leads to a \textit{majority bias}, where the reasoning shifts toward the most frequent patterns in the retrieved cohort.
To address this, the primary objective of Stage 3 is to enforce clinical verification, ensuring that the final recommendation is precisely individualized to the unique needs of the current patient.
The verifier integrates the patient's most recent historical records ($H_t$), the analyzed prescribing tendencies ($T$), and the initial prescription ($y_0$) to ensure clinical appropriateness through a synthesis function $\phi$:
\vspace{-0.5em}
\begin{equation}
    y_{final} = \phi(x_t, H_t, T, y_0)
\vspace{-0.5em}
\end{equation}
To reach a final judgment, the verifier evaluates $y_0$ against 
three refinement decision rules,
treating the preservation of the patient's existing therapy (MAINTAIN) as the baseline for all clinical decisions (see Appendix~\ref{app:prompt_stage3} for detailed 
rules): 
\begin{itemize}
    \item \textbf{Maintain Existing Therapy (MAINTAIN):}  
    Defaulting to preserving the patient's active medication history $H_t$. 
    \item \textbf{Evidence-Supported Addition (ADD):} 
    Adding new drugs only when $T$ provides a high-confidence ADD signal from analogous cases.
    \item \textbf{Evidence-Supported Removal (REMOVE):} 
    Removing drugs only when identifying definitive negative evidence, such as adverse effects or contraindications explicitly documented in the patient's current clinical presentation.
\end{itemize}
Consequently, $y_{final}$ represents the list of refined medications tailored to the patient's current clinical trajectory.

\subsection{Stage 4: Explainable Clinical Summary}
\label{method:stage4}

In the final stage, the model generates an interpretable summary of the verified prescription (prompt provided in Appendix~\ref{app:doctor_summary}). 
The summary generation is formally defined as:
\vspace{-0.5em}
\begin{equation} 
    y_{summary} = \phi(x_t, H, T, y_0, y_{final}) 
\vspace{-0.5em}
\end{equation}
where the input includes the current visit $x_t$, the patient’s longitudinal history $H$, the prescribing tendencies $T$, the initial answer $y_0$, and the finalized prescription $y_{final}$.
The final output, $y_{summary}$, integrates the patient summary, extracted keywords, and supporting clinical evidence alongside the finalized prescription.
For example,
the model details the clinical rationale (e.g., \textit{"Amantadine was maintained to address `weakness in hands',} aligning with the retrieved tendency $T$ and the patient's stable history $H$"), providing a transparent trace of the verification process.

\begin{table*}[t]
\centering
\small
\setlength{\tabcolsep}{3.2pt}
\caption{
\textbf{Evaluation on the HPH and MIMIC-IV.}
We compare task-specific training methods and LLM-based inference-only methods. 
For inference-only methods, \textbf{bold} and \underline{underline} denote the best and second-best performances within each backbone group.
Results are reported as the mean $\pm$ standard deviation across five random seeds.
Statistical significance is computed using a paired t-test (* $p < 0.05$, ** $p < 0.01$).
}
\label{tab:main_results_combined}
\vspace{-2mm}
\resizebox{\textwidth}{!}{
\begin{tabular}{l cccc  cccc}
\toprule
\multirow{2}{*}{\textbf{Method}} 
& \multicolumn{4}{c}{\textbf{HPH}} 
& \multicolumn{4}{c}{\textbf{MIMIC-IV}} \\
\cmidrule(lr){2-5} \cmidrule(lr){6-9}
& \multicolumn{1}{c}{\textbf{F1 score ($\uparrow$)}} 
& \multicolumn{1}{c}{\textbf{Accuracy ($\uparrow$)}} 
& \multicolumn{1}{c}{\textbf{Precision ($\uparrow$)}} 
& \multicolumn{1}{c}{\textbf{Recall ($\uparrow$)}} 
& \multicolumn{1}{c}{\textbf{F1 score ($\uparrow$)}} 
& \multicolumn{1}{c}{\textbf{Accuracy ($\uparrow$)}} 
& \multicolumn{1}{c}{\textbf{Precision ($\uparrow$)}} 
& \multicolumn{1}{c}{\textbf{Recall ($\uparrow$)}} \\
\midrule

Copy-History   & 80.06 & 84.82 & 83.22 & 80.13 & 40.57 & 52.16 & 42.63 & 41.92 \\
\midrule

\multicolumn{9}{c}{\textbf{Task-specific training methods}} \\
\midrule
SafeDrug & -- & -- & -- & -- & 53.80\textcolor{gray}{$\pm$0.00} & 55.68\textcolor{gray}{$\pm$0.00} & 54.67\textcolor{gray}{$\pm$0.00} & 59.06\textcolor{gray}{$\pm$0.00} \\
CogNet   & -- & -- & -- & -- & 35.82\textcolor{gray}{$\pm$0.62} & 32.56\textcolor{gray}{$\pm$0.65} & 28.54\textcolor{gray}{$\pm$0.75} & 83.15\textcolor{gray}{$\pm$0.36} \\
LEADER   & 53.22\textcolor{gray}{$\pm$1.74}
& 58.58\textcolor{gray}{$\pm$1.61}
& 51.52\textcolor{gray}{$\pm$2.10}
& 62.74\textcolor{gray}{$\pm$1.86} & 57.41\textcolor{gray}{$\pm$0.39} & 59.56\textcolor{gray}{$\pm$0.45} & 56.83\textcolor{gray}{$\pm$0.57} & 64.90\textcolor{gray}{$\pm$0.35} \\
Self-RAG 
& 48.72\textcolor{gray}{$\pm$4.55} & 49.85\textcolor{gray}{$\pm$3.40} & 41.84\textcolor{gray}{$\pm$5.71} & 69.85\textcolor{gray}{$\pm$2.02} & 42.98\textcolor{gray}{$\pm$0.22} & 45.22\textcolor{gray}{$\pm$0.26} & 59.04\textcolor{gray}{$\pm$0.19} & 37.09\textcolor{gray}{$\pm$0.21} \\

\midrule
\multicolumn{9}{c}{\textbf{Inference-only methods with large-scale instruct models}} \\
\midrule

Qwen2.5-14B & 61.12\textcolor{gray}{$\pm$1.38}$^{**}$ & 63.18\textcolor{gray}{$\pm$1.79}$^{**}$ & 59.58\textcolor{gray}{$\pm$0.77}$^{**}$ & 69.23\textcolor{gray}{$\pm$1.91}$^{**}$ & 32.24\textcolor{gray}{$\pm$0.22}$^{**}$ & 35.80\textcolor{gray}{$\pm$0.32}$^{**}$ & 50.19\textcolor{gray}{$\pm$0.18}$^{**}$ & 25.75\textcolor{gray}{$\pm$0.23}$^{**}$ \\
Qwen2.5-32B & 78.12\textcolor{gray}{$\pm$0.70}$^{**}$ & 79.94\textcolor{gray}{$\pm$0.83}$^{**}$ & 81.50\textcolor{gray}{$\pm$0.35}$^{**}$ & 79.66\textcolor{gray}{$\pm$0.87}$^{**}$ & 37.19\textcolor{gray}{$\pm$0.33}$^{**}$ & 43.81\textcolor{gray}{$\pm$0.52}$^{**}$ & 50.23\textcolor{gray}{$\pm$0.25}$^{**}$ & 33.02\textcolor{gray}{$\pm$0.47}$^{**}$ \\
Qwen2.5-72B & 72.84\textcolor{gray}{$\pm$0.77}$^{**}$ & 74.61\textcolor{gray}{$\pm$0.96}$^{**}$ & 75.62\textcolor{gray}{$\pm$0.78}$^{**}$ & 75.62\textcolor{gray}{$\pm$0.78}$^{**}$ & 39.17\textcolor{gray}{$\pm$0.20}$^{**}$ & 46.82\textcolor{gray}{$\pm$0.30}$^{**}$ & 48.35\textcolor{gray}{$\pm$0.25}$^{**}$ & 36.76\textcolor{gray}{$\pm$0.24}$^{**}$ \\
\midrule
\multicolumn{9}{c}{\textbf{Inference-only methods with Llama3.1-8B-Instruct}} \\
\midrule
Zero-shot Baseline 
& 49.96\textcolor{gray}{$\pm$0.42}$^{**}$ 
& 53.63\textcolor{gray}{$\pm$0.52}$^{**}$ 
& 47.02\textcolor{gray}{$\pm$0.53}$^{**}$ 
& 60.89\textcolor{gray}{$\pm$0.30}$^{**}$ 
& 24.73\textcolor{gray}{$\pm$0.03}$^{**}$ 
& 24.39\textcolor{gray}{$\pm$0.03}$^{**}$ 
& \underline{49.56}\textcolor{gray}{$\pm$0.12}$^{*}$ 
& 17.58\textcolor{gray}{$\pm$0.03}$^{**}$ \\

Guideline RAG 
& 50.60\textcolor{gray}{$\pm$0.48}$^{**}$ 
& 54.23\textcolor{gray}{$\pm$0.65}$^{**}$ 
& 45.31\textcolor{gray}{$\pm$0.47}$^{**}$ 
& 65.64\textcolor{gray}{$\pm$0.61}$^{**}$ 
& 26.69\textcolor{gray}{$\pm$0.19}$^{**}$ 
& 27.36\textcolor{gray}{$\pm$0.24}$^{**}$ 
& 46.68\textcolor{gray}{$\pm$0.07}$^{**}$ 
& 20.39\textcolor{gray}{$\pm$0.20}$^{**}$ \\

TreatRAG 
& 56.62\textcolor{gray}{$\pm$0.44}$^{**}$ 
& 57.94\textcolor{gray}{$\pm$0.46}$^{**}$ 
& 46.79\textcolor{gray}{$\pm$0.45}$^{**}$ 
& \textbf{86.22}\textcolor{gray}{$\pm$0.62}$^{**}$ 
& \underline{35.28}\textcolor{gray}{$\pm$0.05}$^{**}$ 
& 34.33\textcolor{gray}{$\pm$0.55}$^{**}$ 
& \textbf{65.53}\textcolor{gray}{$\pm$0.14}$^{**}$ 
& 26.07\textcolor{gray}{$\pm$0.05}$^{**}$ \\

MultiQueryRetrieval 
& 30.87\textcolor{gray}{$\pm$1.15}$^{**}$ & 35.70\textcolor{gray}{$\pm$1.14}$^{**}$ & 25.32\textcolor{gray}{$\pm$0.83}$^{**}$ & 47.62\textcolor{gray}{$\pm$2.05}$^{**}$ & 27.56\textcolor{gray}{$\pm$0.33}$^{**}$ & 27.16\textcolor{gray}{$\pm$0.34}$^{**}$ & 50.54\textcolor{gray}{$\pm$0.39}$^{**}$ & 20.24\textcolor{gray}{$\pm$0.28}$^{**}$ \\

MedReflect 
& \underline{62.67}\textcolor{gray}{$\pm$1.46}$^{**}$ 
& \underline{66.76}\textcolor{gray}{$\pm$1.09}$^{**}$ 
& \underline{61.04}\textcolor{gray}{$\pm$0.50}$^{**}$ 
& 71.18\textcolor{gray}{$\pm$2.66}$^{**}$ 
& 28.87\textcolor{gray}{$\pm$0.40}$^{**}$ 
& \underline{36.98}\textcolor{gray}{$\pm$0.28}$^{**}$ 
& 37.09\textcolor{gray}{$\pm$0.69}$^{**}$ 
& \underline{28.15}\textcolor{gray}{$\pm$0.52}$^{**}$ \\

\textbf{PACE-RAG} 
& \textbf{76.30}\textcolor{gray}{$\pm$0.26} 
& \textbf{78.44}\textcolor{gray}{$\pm$0.19} 
& \textbf{76.69}\textcolor{gray}{$\pm$0.41} 
& \underline{80.41}\textcolor{gray}{$\pm$0.33} 
& \textbf{36.93}\textcolor{gray}{$\pm$0.07} 
& \textbf{42.71}\textcolor{gray}{$\pm$0.15} 
& 49.17\textcolor{gray}{$\pm$0.21} 
& \textbf{34.60}\textcolor{gray}{$\pm$0.06} \\

\midrule
\multicolumn{9}{c}{\textbf{Inference-only methods with Qwen3-8B}} \\
\midrule
Zero-shot Baseline 
& \underline{75.57}\textcolor{gray}{$\pm$0.63}$^{**}$ 
& 75.78\textcolor{gray}{$\pm$1.20}$^{**}$ 
& \underline{80.95}\textcolor{gray}{$\pm$0.19}$^{**}$ 
& 75.51\textcolor{gray}{$\pm$0.87}$^{**}$ 
& 34.20\textcolor{gray}{$\pm$0.13}$^{**}$ 
& 38.38\textcolor{gray}{$\pm$0.19}$^{**}$ 
& 47.52\textcolor{gray}{$\pm$0.09}$^{**}$ 
& 29.05\textcolor{gray}{$\pm$0.20}$^{**}$ \\

Guideline RAG 
& 69.67\textcolor{gray}{$\pm$0.50}$^{**}$ 
& 67.03\textcolor{gray}{$\pm$0.60}$^{**}$ 
& 78.26\textcolor{gray}{$\pm$0.35}$^{**}$ 
& 68.28\textcolor{gray}{$\pm$0.54}$^{**}$ 
& 27.51\textcolor{gray}{$\pm$0.03}$^{**}$ 
& 28.48\textcolor{gray}{$\pm$0.07}$^{**}$ 
& 39.95\textcolor{gray}{$\pm$0.07}$^{**}$ 
& 22.58\textcolor{gray}{$\pm$0.05}$^{**}$ \\

TreatRAG 
& 74.86\textcolor{gray}{$\pm$0.30}$^{**}$ 
& 76.29\textcolor{gray}{$\pm$0.29}$^{**}$ 
& 72.42\textcolor{gray}{$\pm$0.45}$^{**}$ 
& \textbf{84.69}\textcolor{gray}{$\pm$0.15}$^{**}$ 
& \underline{41.12}\textcolor{gray}{$\pm$0.21}$^{**}$ 
& 38.93\textcolor{gray}{$\pm$0.32}$^{**}$ 
& \underline{54.76}\textcolor{gray}{$\pm$0.31}$^{**}$ 
& 36.42\textcolor{gray}{$\pm$0.46}$^{**}$ \\

MultiQueryRetrieval 
& 74.87\textcolor{gray}{$\pm$0.12}$^{**}$ & \underline{77.71}\textcolor{gray}{$\pm$0.15}$^{**}$ & 74.78\textcolor{gray}{$\pm$0.21}$^{**}$ & 80.80\textcolor{gray}{$\pm$0.15}$^{**}$ & 36.99\textcolor{gray}{$\pm$0.45}$^{**}$ & 40.78\textcolor{gray}{$\pm$0.76}$^{**}$ & 48.86\textcolor{gray}{$\pm$0.27}$^{**}$ & 32.61\textcolor{gray}{$\pm$0.81}$^{**}$ \\

MedReflect 
& 73.11\textcolor{gray}{$\pm$0.20}$^{**}$ 
& 75.15\textcolor{gray}{$\pm$0.21}$^{**}$ 
& 71.62\textcolor{gray}{$\pm$0.14}$^{**}$ 
& 81.27\textcolor{gray}{$\pm$0.38}$\phantom{^{*}}$ 
& 37.65\textcolor{gray}{$\pm$0.17}$^{**}$ 
& \underline{46.01}\textcolor{gray}{$\pm$0.18}$^{**}$ 
& 42.19\textcolor{gray}{$\pm$0.20}$^{**}$ 
& \underline{38.14}\textcolor{gray}{$\pm$0.21}$^{**}$ \\

\textbf{PACE-RAG} 
& \textbf{80.84}\textcolor{gray}{$\pm$0.12} 
& \textbf{82.75}\textcolor{gray}{$\pm$0.68} 
& \textbf{83.78}\textcolor{gray}{$\pm$0.32} 
& \underline{81.90}\textcolor{gray}{$\pm$1.22} 
& \textbf{47.22}\textcolor{gray}{$\pm$0.12} 
& \textbf{52.32}\textcolor{gray}{$\pm$0.17} 
& \textbf{57.03}\textcolor{gray}{$\pm$0.05} 
& \textbf{45.94}\textcolor{gray}{$\pm$0.17} \\

\bottomrule
\end{tabular}
}
\vspace{-1em}
\end{table*}

\section{Experimental Setup}
\vspace{-1mm}
\subsection{Datasets}
\label{sec:datasets}
We evaluate PACE-RAG on two clinical cohorts: a real-world unstructured Parkinson's Disease cohort from HPH and the structured MIMIC-IV benchmark~\cite{johnson_2023_mimic_iv}.

\paragraph{Real-world unstructured cohort (HPH).} 
We constructed a large-scale longitudinal dataset derived from raw clinical notes at Inje University Haeundae Paik Hospital (HPH), Korea. 
This cohort comprises clinical data from 947 patients diagnosed with Parkinson's Disease between March 2010 and September 2023. We gathered SOAP (Subjective, Objective, Assessment, Plan) notes from both initial and follow-up visits using Electronic Medical Records (EMRs). To ensure patient privacy, all personally identifiable information was anonymized. Detailed dataset specifications are provided in Appendix~\ref{app:data_preprocess_hph}.

\paragraph{Benchmark cohort (MIMIC-IV).} 
To evaluate PACE-RAG in a substantially different structured clinical setting,
we utilized the MIMIC-IV dataset~\cite{johnson_2023_mimic_iv}, a standard benchmark for EMRs. We utilized clinical data from a subset of 196,501 patients within the \texttt{hosp} module, specifically utilizing the \texttt{admissions}, \texttt{diagnoses\_icd}, \texttt{d\_icd\_diagnoses}, and \texttt{prescriptions} tables. 
To standardize medication naming, we mapped raw National Drug Codes (NDC) to their corresponding Anatomical Therapeutic Chemical (ATC) classification codes. 
Details of the MIMIC-IV data processing are provided in Appendix~\ref{app:data_preprocess_mimic}.

\subsection{Retrieval Pool and Test Set Splitting}
\label{sec:data_split}
To prevent data leakage, we implemented a patient-level splitting strategy across both cohorts.
By partitioning the data based on unique Patient IDs, we strictly ensured that no overlap occurred between the patients in the retrieval pool and those in the test set. 
For the HPH dataset, records were divided into a retrieval pool and a test set following a strict 80:20 ratio. 
For the MIMIC-IV dataset, we sampled the test set to approximately 1,000 admissions, with the remaining records constituting the retrieval pool.
This configuration ensures a fair comparison and aligns our experimental setup with the baseline~\cite{treatrag_2025}.


\subsection{Baselines}

To evaluate PACE-RAG, we compare it with task-specific training methods, standard RAG baselines, and verification-based methods using Qwen3-8B~\cite{qwen3technicalreport} and Llama3.1-8B-Instruct~\cite{llama3modelcard} as backbone models.
Detailed implementation settings are provided in Appendix~\ref{app:baselines_detail}.

\paragraph{Copy-History:}
Copy-History directly uses the medication list from the patient's most recent prior visit as the current recommendation.
\paragraph{Task-specific training methods:}
SafeDrug~\cite{yang2021safedrug} and CogNet~\cite{wu2022cognet} are non-LLM neural medication recommendation methods trained on structured patient records.
LEADER~\cite{liu2024leader} is a LLM-based medication recommendation method that requires task-specific training.
Self-RAG~\cite{asai2024selfrag} is a learning-based RAG baseline.

\paragraph{Inference-only baselines:}
The Zero-shot baseline generates prescriptions from the current clinical note $x_t$ and patient history $H_t$ without retrieval.
We also evaluate larger instruct models under the same zero-shot setting.
For RAG baselines, Guideline RAG retrieves clinical guideline passages, TreatRAG~\cite{treatrag_2025} retrieves similar historical patient cases, and MultiQueryRetrieval~\cite{langchain_multiqueryretriever} expands the query into multiple variants.
MedReflect~\cite{huang_2026_medreflect} refines prescription through self-generated reflective question--answer pairs.

\section{Experimental Results}
\label{sec:results}
\vspace{-1mm}


\begin{figure}[t]
    \centering

    \includegraphics[width=1\linewidth]{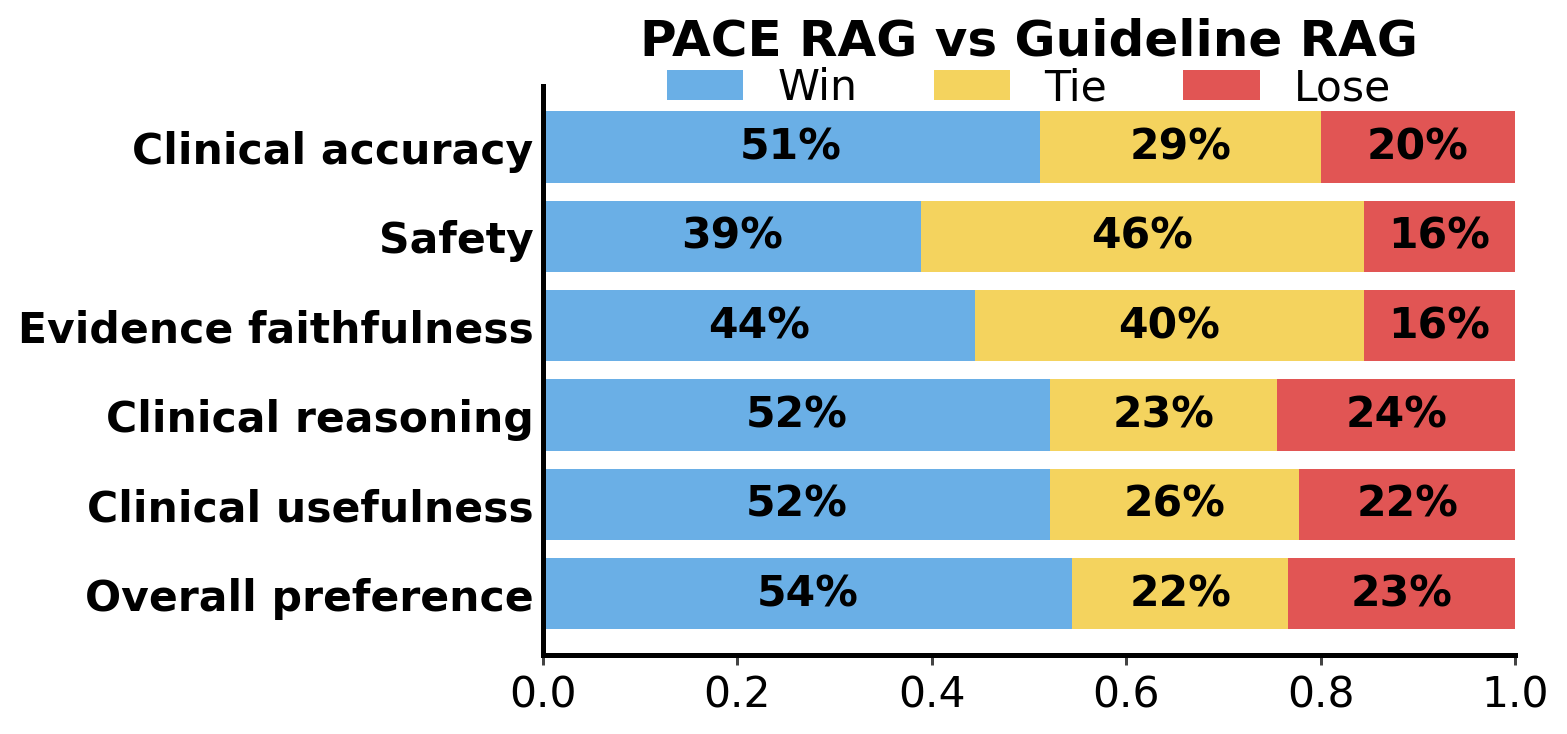}

    \vspace{0.5em}

    \includegraphics[width=1\linewidth]{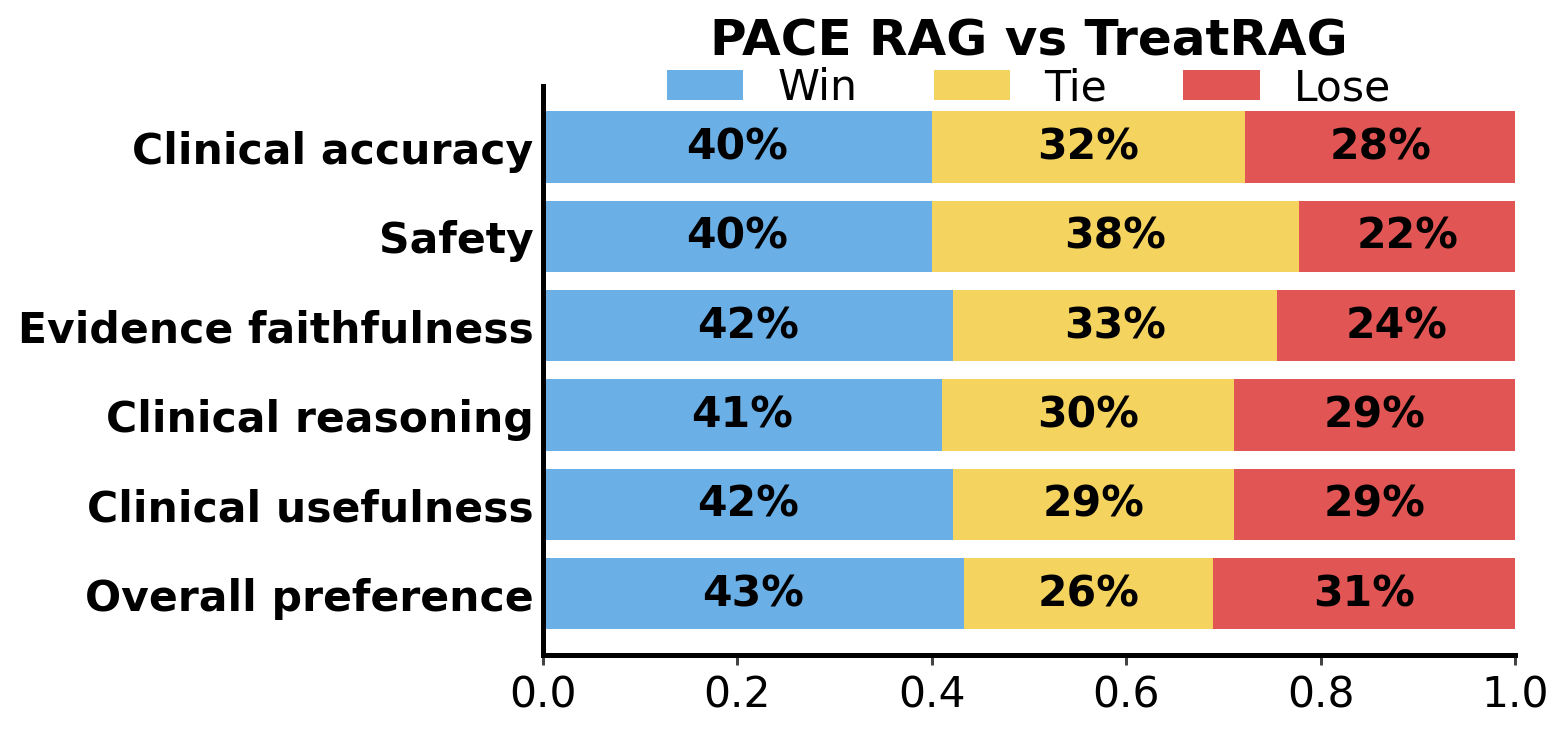}

    \caption{
    \textbf{Specialist evaluation results.}
    Comparative assessment of PACE-RAG against Guideline RAG (top) and TreatRAG (bottom) by clinical specialists across six evaluation criteria.}
    \label{fig:specialist_results}

    \vspace{-1em}
\end{figure}

\paragraph{Real-world unstructured data results.}
\label{subsec:main_results}
Table~\ref{tab:main_results_combined} (Left)
presents a comparative performance analysis on the HPH dataset, which comprises real-world unstructured clinical narratives. 
Task-specific training methods such as SafeDrug and CogNet are designed for structured inputs and therefore are not directly applicable to this dataset.
In contrast, PACE-RAG can operate directly on unstructured clinical notes and achieves strong performance without task-specific training.
With Llama3.1-8B-Instruct, PACE-RAG obtains the highest F1 score of 76.30\% and accuracy of 78.44\% among inference-only LLM-based methods.
This trend is consistent with Qwen3-8B, where PACE-RAG achieves an F1 score of 80.84\% and an accuracy of 82.75\%.
Compared with other RAG or verification-based methods, PACE-RAG also provides a more balanced precision--recall trade-off, whereas methods such as TreatRAG tend to over-predict positive cases with high recall but lower precision.
Additional precision--recall trade-off analyses and ablation studies on hyperparameter robustness are provided in Appendix~\ref{app:precision-recall-tradeoff-results} and ~\ref{app:hyperparamter_sensitivity}.

\begin{table*}[!t]
\small
\centering

\begin{minipage}[t]{0.4\textwidth}
    \vspace{0pt}
    \centering
    \setlength{\tabcolsep}{2.2pt}
    \renewcommand{\arraystretch}{1.05}
    \captionof{table}{
    \textbf{Medication recommendation and safety metrics.}
    Higher Jaccard similarity and lower DDI rate indicate better performance.
    }
    \label{tab:medication_metrics}
    \vspace{-2mm}
    \resizebox{\linewidth}{!}{
    \begin{tabular}{lcc cc}
    \toprule
    \multirow{2}{*}{\textbf{Method}} 
    & \multicolumn{2}{c}{\textbf{MIMIC-IV}} 
    & \multicolumn{2}{c}{\textbf{HPH}} \\
    \cmidrule(lr){2-3} \cmidrule(lr){4-5}
    & \textbf{Jaccard ($\uparrow$)} 
    & \textbf{DDI ($\downarrow$)} 
    & \textbf{Jaccard ($\uparrow$)} 
    & \textbf{DDI ($\downarrow$)} \\
    \midrule
    Ground Truth & --     & 0.0696 & --     & 0.2359 \\
    Zero-shot    & 0.1704 & \textbf{0.0535} & 0.6870 & 0.2624 \\
    TreatRAG     & 0.2105 & 0.0566 & 0.6746 & 0.2360 \\
    \textbf{PACE-RAG} 
                  & \textbf{0.2540} & 0.0635 
                  & \textbf{0.7664} & \textbf{0.2300} \\
    \bottomrule
    \end{tabular}
    }
\end{minipage}
\hfill
\begin{minipage}[t]{0.32\textwidth}
    \vspace{0pt}
    \centering
    \setlength{\tabcolsep}{2.5pt}
    \renewcommand{\arraystretch}{1.08}
    \captionof{table}{
    \textbf{Performance comparison of different verification methods.}
    Our method outperforms the no-verification baseline and MedReflect.
    }
    \label{tab:effect_of_verifier}
    \vspace{-2mm}
    \begin{tabular}{l|cc}
    \toprule
    Method & F1 & Acc \\
    \midrule
    Stage 1 & 0.7665 & 0.7858 \\
    + MedReflect & 0.7268 & 0.7088 \\
    \textbf{+ Ours (Stage 2\&3)} & \textbf{0.8101} & \textbf{0.8322} \\
    \bottomrule
    \end{tabular}
\end{minipage}
\hfill
\begin{minipage}[t]{0.26\textwidth}
    \vspace{0pt}
    \centering
    \setlength{\tabcolsep}{2.5pt}
    \renewcommand{\arraystretch}{1.08}
    \captionof{table}{
        \textbf{Effect of different retrieval sources.} 
    We compare retrieval from guidelines, similar patients, and both.
    }
    \label{tab:ours_with_guideline}
    \vspace{-2mm}
    \begin{tabular}{l|cc}
    \toprule
    RAG docs. & F1 & Acc \\
    \midrule
    Guideline & 0.6941 & 0.6681 \\
    Sim. Patients & 0.8101 & 0.8322 \\
    Both & 0.8009 & 0.8257 \\
    \bottomrule
    \end{tabular}
\end{minipage}

\vspace{-4mm}
\end{table*}

\paragraph{Evaluation on a structured clinical benchmark.}
\label{subsec:mimic_results}
To evaluate whether PACE-RAG remains effective outside the HPH setting,
we further apply PACE-RAG to the large-scale MIMIC-IV benchmark
(Table~\ref{tab:main_results_combined}, Right).
On MIMIC-IV, task-specific training methods can achieve higher performance because they are explicitly optimized for structured medication recommendation.
Nevertheless, PACE-RAG remains highly competitive among inference-only LLM-based methods.
With Llama3.1-8B-Instruct, PACE-RAG achieves the best F1 score of 36.93\% and accuracy of 42.71\% among the LLM-based baselines.
With Qwen3-8B, PACE-RAG further improves performance and achieves the best results across all metrics among inference-only methods.

\paragraph{Comparison with large-scale models.}
\label{subsec:results_large_model_mimic}
Furthermore, we evaluated our framework against models with significantly larger parameter capacities
(Table~\ref{tab:main_results_combined}).
We conducted a comparative analysis using local, open-source models, including Qwen2.5-14B, 32B, and 72B.
PACE-RAG with Qwen3-8B consistently outperforms these significantly larger dense models.
This finding suggests that a structured inference pipeline can successfully compensate for the advantages usually gained by increasing model capacity.
These results demonstrate that PACE-RAG enables efficient on-premise clinical deployment, offering a cost-effective alternative to large-scale models.

\paragraph{Specialist evaluation.}
\label{subsec:human_eval}
To assess the reasoning capabilities of PACE-RAG, we performed a blinded human evaluation with three Parkinson’s disease specialists. Using the HPH dataset, we provided model-generated responses to the experts to examine how closely each model’s reasoning process resembles that of experienced clinicians. The specialists conducted a comparative evaluation of PACE-RAG and the Guideline RAG across six criteria: Clinical Accuracy, Safety, Faithfulness, Clinical Reasoning, Clinical Usefulness, and Overall Preference. 
For each comparison, evaluators selected the preferred response (Win) or indicated that both responses were comparable (Tie).
Detailed descriptions of each evaluation criterion are provided in Appendix~\ref{app:human_eval_protocol}.

As illustrated in Figure~\ref{fig:specialist_results}, which compares PACE-RAG with Guideline RAG (top) and TreatRAG (bottom), 
PACE-RAG was consistently favored over Guideline RAG, achieving a 76\% Win/Tie rate in Overall Preference and Win/Tie rates above 70\% across all other clinical dimensions.
Against the stronger TreatRAG baseline, PACE-RAG was also preferred more frequently across the evaluated criteria.
In Overall Preference, PACE-RAG was selected in 43\% of cases, compared with 31\% for TreatRAG, with the remaining 26\% resulting in ties.
These results suggest that the improvements of PACE-RAG extend beyond prescription-overlap metrics, with clinical specialists generally preferring its recommendations and supporting rationales over those produced by the comparison methods.

\section{Discussions}
\label{sec:analysis}
\vspace{-1mm}

\begin{wrapfigure}{r}{0.55\linewidth}
    \vspace{-2.5mm}
    \centering
    \includegraphics[width=\linewidth]{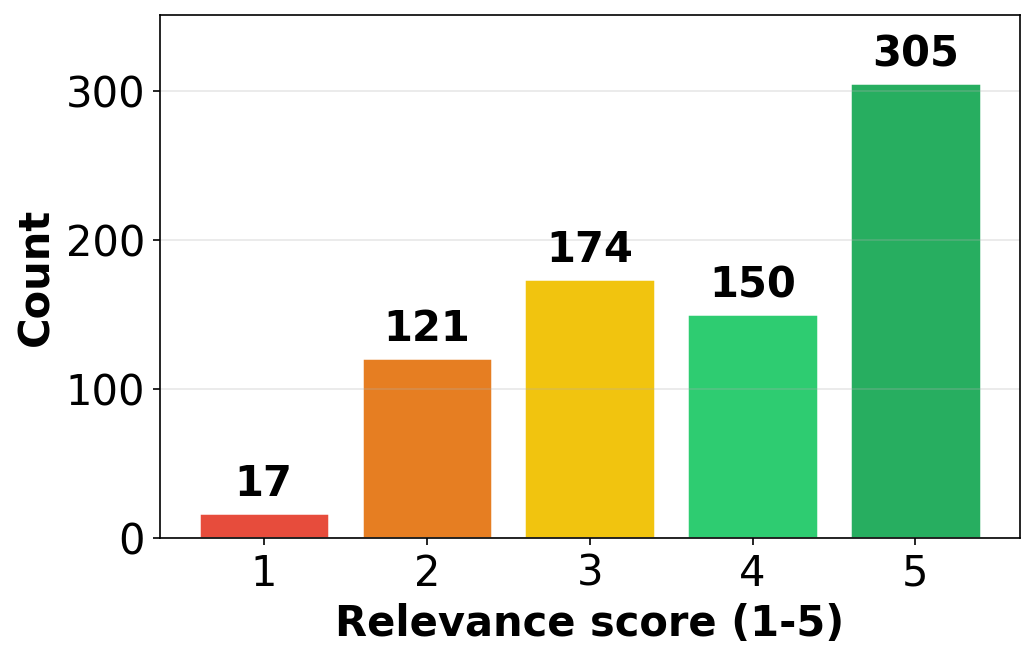}
    \caption{
    \textbf{Relevance scores.}
    The distribution of relevance scores between the extracted keywords and the ground truth
    (\text{Mean}=$3.79$, \text{SD}=$1.19$).
    }
    \label{fig:gpt4-score-of-extractor}
    \vspace{-3.5mm}
\end{wrapfigure}

\paragraph{Are the extracted keywords representative of patient symptoms?}
\label{subsec:effect_of_focus query}
To evaluate the contribution of Stage 1,
we examine whether the extracted keywords are clinically relevant 
by measuring 
their alignment with the medications actually prescribed to patients. 
As shown in Figure~\ref{fig:gpt4-score-of-extractor}, 
GPT-4o is used as an external evaluator to score the relevance between the extracted keywords and the prescribed medications on the HPH dataset. 
More than half of the cases receive high relevance scores (4–5), indicating that the extracted 
focus queries 
$\{f_1, ..., f_i\}$
capture clinically meaningful signals related to treatment decisions.
This makes retrieval patient-specific because PACE-RAG retrieves evidence around the clinical features extracted from the current patient, rather than globally similar records based on the entire note.
The detailed evaluation prompt is provided in Appendix Figure~\ref{fig:prompt_keyword_evaluator}.

\paragraph{Does PACE-RAG improve both prescription alignment and safety?}
We further evaluate medication recommendation quality using Jaccard similarity and Drug--Drug Interaction (DDI) rate.
Jaccard similarity measures prescription-set overlap with the ground truth, while DDI rate measures the proportion of potentially harmful drug pairs.
As shown in Table~\ref{tab:medication_metrics}, PACE-RAG achieves the highest Jaccard similarity on both MIMIC-IV and HPH, indicating better alignment with physician prescriptions.
Although PACE-RAG slightly increases the DDI rate over Zero-shot and TreatRAG on MIMIC-IV, it remains below the ground-truth DDI rate.
On HPH, PACE-RAG achieves the lowest DDI rate among all methods, including the ground truth.
These results suggest that PACE-RAG improves prescription-set alignment while maintaining a comparable or lower medication safety risk.

\paragraph{Is tendency analysis and refinement more robust than internal self-reflection?}
\label{subsec:effect_of_tendency_verification}
To assess our verification stage, we compared the Stage 1 baseline against two strategies: the MedReflect method and our proposed Stage 2\&3 modules.
Table~\ref{tab:effect_of_verifier} 
reveals that incorporating MedReflect leads to a performance drop compared with Stage 1 alone.
This degradation suggests that the model becomes overly sensitive to the context of its own generated question-answer pairs.
In contrast, the integration of our proposed Stage 2\&3 modules yields a substantial improvement in both F1 score and accuracy. 
These results demonstrate that the integration of Stages 2\&3 is essential for the verification of retrieved data, elevating the overall performance and reliability of the generated prescriptions.

\paragraph{Does combining clinical guidelines with patient cases provide synergistic gains?}
\label{subsec:guideline_with_ours}
We further investigate the impact of different retrieval sources by incorporating clinical guidelines alongside similar patient cases. 
As detailed in Table~\ref{tab:ours_with_guideline}, retrieving similar patient cases alone yields the highest performance, with an F1 score of 81.01\% and an accuracy of 83.22\%. 
Interestingly, integrating both sources does not provide additional gains.
This suggests that the inclusion of long clinical guidelines may introduce unnecessary complexity or noise, potentially distracting the model during reasoning, as also noted in prior work~\cite{chen2026medcopilotmedicalassistantpowered}.

\begin{figure}[t]
\centering

\begin{minipage}[t]{0.52\linewidth}
\centering
\small
\setlength{\tabcolsep}{5pt}
\renewcommand{\arraystretch}{1.15}
\captionof{table}{
\textbf{Computational cost and latency analysis.}
We compare the per-case inference latency of Zero-shot and PACE-RAG.
}
\label{tab:latency_analysis}
\resizebox{\linewidth}{!}{
\begin{tabular}{lcc}
\toprule
\textbf{Metric} 
& \textbf{Zero-shot}
& \textbf{PACE-RAG} \\
\midrule
Avg. time/case (s) & 0.845 & 3.047 \\
Median (s)         & 0.842 & 2.808 \\
Relative latency   & 1.00$\times$ & 3.61$\times$ \\
\bottomrule
\end{tabular}
}
\end{minipage}
\hfill
\begin{minipage}[t]{0.44\linewidth}
\centering
\vspace{-2.5mm}
\includegraphics[width=\linewidth]{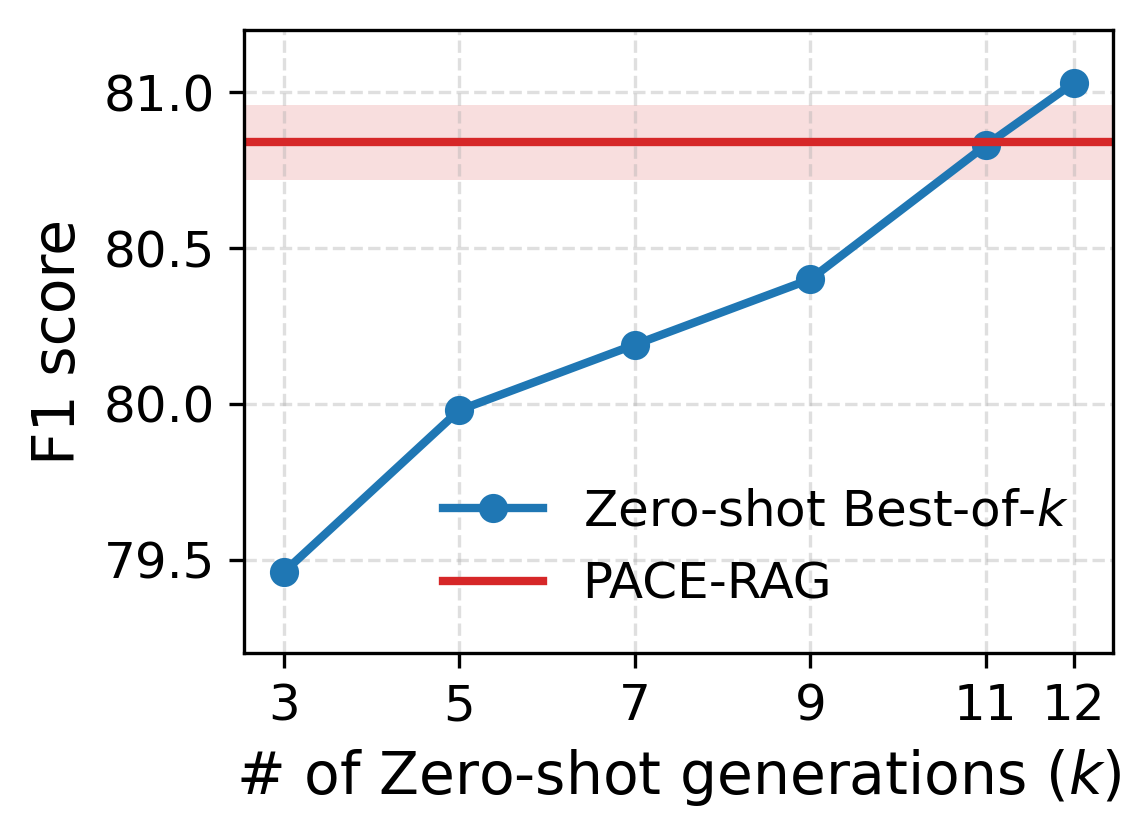}
\caption{
\textbf{Best-of-$k$ performance of repeated Zero-shot generations.}
}
\label{fig:best_of_k_latency}
\end{minipage}
\vspace{-2em}
\end{figure}

\paragraph{What is the trade-off between inference cost and prescription quality?}
\label{subsection:latency}

We analyze the computational cost and latency of PACE-RAG.
As shown in Table~\ref{tab:latency_analysis}, PACE-RAG requires higher inference time than the single-call Zero-shot baseline because it consists of multiple sequential LLM calls.
Using Zero-shot as the latency reference, PACE-RAG takes 3.61$\times$ longer per case.

To contextualize this cost, we additionally compare PACE-RAG with repeated Zero-shot generation under a Best-of-$k$ F1 setting.
Here, Best-of-$k$ F1 denotes the highest F1 score among $k$ independently generated Zero-shot prescription sets for each case.
As shown in Figure~\ref{fig:best_of_k_latency}, the Zero-shot baseline requires approximately 12 repeated generations to reach performance comparable to PACE-RAG.
Thus, although PACE-RAG is slower than a single Zero-shot generation, it achieves comparable performance with fewer inference attempts than repeated end-to-end Zero-shot sampling.

\section{Conclusion}
\label{sec:conclusion}
In this work, we proposed PACE-RAG, a four-stage pipeline designed to produce patient-specific drug recommendations by explicitly reasoning over the key symptoms of each patient.
Across a real-world Parkinson’s disease cohort and the MIMIC-IV benchmark, PACE-RAG consistently produced more 
patient-specific
and clinically evidence-based recommendations than other RAG methods and verification baselines, 
showing consistent performance across both unstructured and structured clinical settings.
Moreover, blinded qualitative evaluation by Parkinson’s specialists indicated that our approach generates reasoning that is more clinically meaningful.
We anticipate that this framework can be further expanded to clinical systems that integrate multi-modal clinical signals—such as laboratory results, imaging, and structured risk factors—
while evaluating robustness across diverse institutional prescribing styles.

\section*{Limitations}
We acknowledge several limitations in our study.

\begin{itemize}
    \item \textbf{Generalizability and Single-center Bias:} 
    Our evaluation of unstructured clinical narratives is limited to the single-center HPH Parkinson's disease cohort. Although MIMIC-IV provides evaluation in a substantially different clinical setting, it is based on structured diagnostic and medication records and therefore does not establish robustness to unstructured documentation styles across institutions. Future work should validate PACE-RAG on multi-center unstructured EMRs with diverse clinical documentation practices.
    
    \item \textbf{Computational Latency and Inference Cost:} The proposed four-stage pipeline requires multiple sequential LLM inferences to reach a final prescription. While this modular approach enhances performance, it increases the total inference time and computational overhead compared to zero-shot baseline models. 
    This additional latency may limit real-time use in time-sensitive clinical environments, motivating future work on parallelized or more efficient inference.
    
    \item \textbf{Prompt Engineering and Human Effort:} Each stage of the PACE-RAG pipeline necessitates task-specific prompts to ensure optimal performance. This process currently relies on manual engineering, which limits the framework's immediate scalability to new clinical domains. 
    Future work could reduce this manual effort through automatic prompt adaptation or agent-based prompt optimization when transferring the framework to new clinical domains.
    
    \item \textbf{Linguistic Nuance Loss in Translation:} For the HPH dataset, clinical records were translated from Korean to English for processing. This translation step may have inadvertently missed subtle linguistic nuances inherent in the source language, potentially affecting the depth of diagnostic reasoning. 
    We did not evaluate alternative translation methods in the current study because institutional data-governance constraints prevented the use of external proprietary LLM APIs for these clinical records. Future work will explore institution-approved or secure on-premise medical translation systems to better preserve clinical nuance while maintaining data privacy.
\end{itemize}
Because incorrect medication recommendations may lead to inappropriate treatment or adverse drug events, PACE-RAG is intended as a research prototype rather than an autonomous clinical decision-making system. Any real-world deployment would require prospective clinical validation and oversight by qualified healthcare professionals.

\section*{Ethics Statement}
This retrospective study was approved by the Institutional Review Board (IRB No. 2023-11-042), which waived the requirement for informed consent because the study used fully de-identified data. 
In addition, this work utilizes the MIMIC-IV database in accordance with its approved data-use agreement.
Due to privacy restrictions on the institutional HPH cohort, we provide code and scripts for the MIMIC-IV evaluation.

\section*{Acknowledgements}
This research was supported by the National Research Foundation of Korea (NRF) under Grants RS-2023-00262527 and RS-2024-00336454.

\paragraph{AI Assistance.} 
Generative AI assistants were used for language editing and coding assistance. All scientific content and generated code were reviewed and verified by the authors. 
No private or patient-identifiable clinical information was provided to external AI assistants.

\newpage
\bibliography{custom}

\newpage
\appendix

\section{Data Details}
\label{app:data_detail}

\subsection{Real-World Unstructured Cohort (HPH)}
\label{app:data_preprocess_hph}

\paragraph{Data Statistics.}

\begin{table}[h]
\centering
\small
\caption{\textbf{Real-world unstructured cohort statistics (HPH).} Patient counts and visit counts before and after exclusion, 
with patient-level splitting into retrieval pool and test sets.}
\label{tab:appendix_HPH_stats}
\adjustbox{max width=\columnwidth}{
\begin{tabular}{lrr}
\toprule
\textbf{Item} & \textbf{\#Patients} & \textbf{\#Visits} \\
\midrule
Original cohort (raw) & 947 & -- \\
Excluded: w/o visit records & 617 & -- \\
\midrule
Total & 330 & 4{,}216 \\
Retrieval pool split (patient-level) & 266 & 3{,}449 \\
Test split (patient-level) & 64 & 767 \\
\bottomrule
\end{tabular}
}
\vspace{-0.3cm}
\end{table}

From an initial pool of 947 Parkinson's disease patients, we applied strict filtering criteria to ensure longitudinal integrity, excluding individuals who lacked either initial or follow-up visit records. 
As shown in Table \ref{tab:appendix_HPH_stats}, this process resulted in a refined dataset of 330 patients and 4,216 total visits.
Following an 80:20 patient-level split, the final dataset comprises 266 patients in the historical patient database (3,449 total visits for retrieval) and 64 patients in the evaluation set (767 total visits for testing).

\paragraph{Translation.}

Since the original clinical records were documented in Korean, we translated the entire dataset into English using the Google Translator API. 

\paragraph{Data Features.}
We utilized clinical notes from both initial and follow-up visits. 
Although the original records were unstructured, they contained explicit tags for the Subjective (S), Objective (O), Assessment (A), and Plan (P) components. To transform these into a structured format, we employed regular expressions to parse each segment, ensuring a consistent SOAP framework across the dataset.
To maintain the model's focus on clinical narratives, all non-textual information outside the SOAP structure was excluded. 

For our experimental task, while each historical record $x_i \in H$ contains the full SOAP components, the current note $x_t$ specifically excludes the `Plan' (P) part. Our objective is to generate this missing `Plan' by determining a set of multiple drugs $Y = \{y_1, y_2, \dots, y_n\}$ ($n \geq 1$). 
\paragraph{Number of Medications.} 
The HPH dataset contains 31 unique medications in the retrieval pool and 24 in the test set. 
The detailed lists and counts of these medications for each split are summarized in Table \ref{tab:medications_list}.

\subsection{Benchmark Cohort (MIMIC-IV)}
\label{app:data_preprocess_mimic}

\begin{table}[h]
\centering
\small
\caption{\textbf{Benchmark cohort statistics (MIMIC).} From the full dataset, patients were randomly sampled at the \texttt{subject\_id}, and the selected subset was split into retrieval pool and test sets at the patient level.}
\label{tab:appendix_mimic_stats}
\adjustbox{max width=\columnwidth}{
\begin{tabular}{lrr}
\toprule
\textbf{Item} & \textbf{\#Patients} & \textbf{\#Admissions} \\
\midrule
Original cohort (raw) & 364{,}627 & 546{,}028 \\
Excluded: mapping error & 168{,}126 & 83{,}295 \\
\midrule
Total & 196{,}501 & 462{,}733 \\
Retrieval pool split (patient-level) & 196{,}110 & 461{,}639 \\
Test split (patient-level) & 391 & 1{,}094 \\
\bottomrule
\end{tabular}
}
\vspace{-0.3cm}
\end{table}

\paragraph{Data Statistics.}

For the MIMIC-IV benchmark, the dataset was partitioned  by \texttt{subject\_id} into a comprehensive retrieval pool and a dedicated evaluation set.
As detailed in Table \ref{tab:appendix_mimic_stats}, the retrieval pool serves as a massive historical patient database, comprising 196,110 unique patients and their corresponding 461,639 hospital admissions. In contrast, the evaluation set consists of 391 unique patients across 1,094 admissions.

\paragraph{Data Features.}
For each patient, the input consists of a historical sequence of admissions $H = \{x_1, x_2, \dots, x_{t-1}\}$ and the diagnostic context of the current admission $x_t$. Each admission record is identified by its \texttt{hadm\_id} from the \texttt{hosp} module. To provide explicit semantic context for the model, raw ICD-9 and ICD-10 codes from the \texttt{diagnoses\_icd} table are mapped to their corresponding descriptive text representations. 
Consequently, each historical record $x_i$ consists of a comprehensive list of diagnosis names associated with that specific visit.

We denote the set of prescribed medications as $Y = \{y_1, y_2, \dots, y_n\}$ ($n \geq 1$), retrieved from the \texttt{prescriptions} table.
To achieve clinical standardization, raw National Drug Codes (NDC) are mapped to the Anatomical Therapeutic Chemical (ATC) classification system via the RxNorm API. During this process, prescriptions that failed to map to a valid ATC code (e.g., non-pharmacological noise such as normal saline or medical supplies) were strictly excluded to ensure dataset quality.

\paragraph{Number of Medications.}
The MIMIC dataset encompasses a substantially broader range of treatments, comprising 406 unique medications in the retrieval pool and 254 in the test set. Detailed lists are summarized in Table \ref{tab:medications_list}.

\section{Method Details}
\label{app:method_detail}

\subsection{Hyperparameters}
\label{app:hyperparamter}

\begin{table}[h]
\centering
\small
\caption{\textbf{Hyperparameters used in PACE-RAG experiments.}}

\label{tab:hyperparameters}
\adjustbox{max width=\columnwidth}{
\begin{tabular}{lcc}
\toprule
\textbf{Parameter} & \textbf{LLaMA} & \textbf{Qwen} \\
\midrule
\multicolumn{3}{l}{\textit{Retrieval Settings}} \\
Retrieval top-$k$ & 7 & 7 \\
Threshold $\tau$ & 0.9 & 0.9 \\
\midrule
\multicolumn{3}{l}{\textit{Generation Settings}} \\
Context length (\texttt{num\_ctx}) & 8192 & 4096 \\
Max token length (\texttt{num\_predict}) & 400 & 220 \\
Temperature & 0.8 & 0.6 \\
\bottomrule
\end{tabular}
}
\vspace{-0.3cm}
\end{table}
Table~\ref{tab:hyperparameters} summarizes the key retrieval and generation hyperparameters used in our experiments.
Detailed ablation studies regarding the retrieval configurations are further explored in Appendix~\ref{app:ablation}~(Figure~\ref{fig:performance_hyperparameter}).
We used five fixed random seeds for the experiments: 42, 137, 2025, 3141, 7777.

\subsection{Details of Extraction Focus Query}
\label{app:data_extract_focus}

We employ a \textbf{Focus Query Extractor} to extract key symptoms from the clinical note $x_t$. The specific prompt utilized for this extraction process is provided in Appendix \ref{app:prompts_focus_query_extractor}.
To ensure the high fidelity and clinical relevance of the extracted information, the extractor adheres to two primary operational constraints:
\begin{enumerate}
    \item \textbf{Exact substring constraint.}
    The model is restricted to extracting only literal, exact substrings directly from the original clinical note. 
    \item \textbf{Actionable clinical signal filtering.}
    The extractor prioritizes actionable clinical signals. 
    If the patient is clinically stable, showing improvement, or if the note contains only conversational remarks without acute issues, the model returns an empty list, ensuring that the RAG pipeline is triggered only when significant clinical indicators require a personalized prescribing strategy.
\end{enumerate}

\paragraph{Real-World Unstructured Cohort (HPH).}

For the real-world hospital dataset, focus queries $\{f_1, f_2, ..., f_n\}$ are extracted specifically from the `Subjective' and `Assessment' parts of the clinical notes. 
To minimize LLM confusion and prevent the generation of overly broad search terms that degrade retrieval quality (e.g., extracting generic terms like ``Parkinson's disease'' or ``worse''), we constrain the extractor to output a maximum of two highly specific, contextual symptom phrases per patient.

\paragraph{Benchmark Cohort (MIMIC-IV).}

Unlike the Parkinson's disease dataset, we extract up to two primary pathological conditions (e.g., ``Acute kidney failure'', ``Sepsis'') directly from the parsed list of diagnosis titles. 
To avoid ambiguity, 
the model is strictly prohibited from omitting essential descriptors (e.g., `acute' or `chronic') or outputting vague, single-word terms like `failure' alone.

\subsection{Implementation of Patient Similarity Search}
\label{app:rag_detail}

We utilize a Vector Database (FAISS) to identify similar cases from the historical pool $D$. 
Each record is selected based on its semantic similarity to the clinical focus $f_i$, where embeddings are generated using SentenceTransformer (\texttt{all-MiniLM-L6-v2})~\cite{reimers-2019-sentence-bert}. 
Retrieval is performed with a similarity threshold $\tau = 0.9$ and a maximum of $k = 7$ records per focus query.

\paragraph{Real-World Unstructured Cohort (HPH).}
We retrieve a historical database $D = \{d_1, d_2, \dots, d_N\}$, where each record $d_j$ represents the latest SOAP-formatted clinical note for each patient.
Specifically, queries $f_i$ derived from the Subjective part are used to retrieve Subjective part from the database, while those derived from the Assessment part are mapped to analogous Assessment part.

\paragraph{Benchmark Cohort (MIMIC-IV).}
In the MIMIC-IV benchmark, the historical database $D = \{d_1, d_2, \dots, d_N\}$ consists of records where each entry $d_j$ is represented the most recent clinical status, formatted as a comma-separated list of pathological conditions. 
For each focus query $f_i$, we retrieve the historical admissions from $D$ that exhibit the highest semantic similarity to the query.

\subsection{Baseline Implementation Details}
\label{app:baselines_detail}
\paragraph{SafeDrug:}
SafeDrug~\cite{yang2021safedrug} is a task-specific medication recommendation model that predicts a medication set from longitudinal diagnosis and procedure sequences while explicitly controlling drug--drug interaction (DDI) risk.
We adapt our MIMIC-IV cohort to SafeDrug's structured input format by retaining admissions with diagnosis, procedure, and medication information, mapping medications to ATC codes, and constructing patient-level visit sequences with at least two medication visits.
For evaluation, we keep only test visits compatible with the training vocabulary and remove visits with empty diagnosis, procedure, or medication fields.
The model is trained for 50 epochs, and validation DDI rate, Jaccard similarity, PRAUC, precision, recall, F1 score, and average medication count are monitored at each epoch.

\paragraph{COGNet:}
COGNet~\cite{wu2022cognet} is a graph-augmented sequence-to-sequence medication recommendation model that generates a drug set autoregressively using a copy-or-predict mechanism over cross-visit diagnosis, procedure, and medication 
histories. We adapt our MIMIC-IV cohort to COGNet's structured input format by mapping NDC codes to ATC-3 via the standard NDC--RXCUI--ATC pipeline, retaining the top-300 most frequent ATC-3 codes, and constructing patient-level visit sequences with at least two qualifying visits. The DDI adjacency matrix is constructed from the bio-Decagon polypharmacy side-effect dataset, and the EHR co-occurrence matrix is built from training records. The train/test split follows our cohort-level patient split, ensuring no overlap between training and test subjects. The model is trained for 50 epochs with Adam (lr~$=10^{-4}$, batch size~$=16$, embedding dimension~$=64$), and performance is evaluated via bootstrapping over 10 random 80\% subsamples of the test set.
\paragraph{LEADER:}
LEADER~\cite{liu2024leader} is trained with a two-stage pipeline under a unified setting for fair baseline comparison. In Stage 1, the model learns medication recommendation from longitudinal EHR visits by encoding diagnosis/procedure/medication sets with Transformer blocks, aggregating visit-level trajectories, and predicting multi-label medications via a BCE-based objective. In Stage 2, the Stage-1 checkpoint is used as initialization and optimized with feature-based knowledge distillation from a LLaMA 3.1-8B teacher, where the student intermediate representation is projected to the teacher hidden-state space and trained with an MSE distillation loss, i.e., \(\mathcal{L}=\mathcal{L}_{\mathrm{BCE}}+\alpha\mathcal{L}_{\mathrm{KD}}\). To align with other baselines, we set the teacher backbone to LLaMA 3.1-8B across experiments and use the same preprocessing pipeline as COGNet for MIMIC. The model is trained for up to 50 epochs with Adam (\(\mathrm{lr}=5\times10^{-4}\), batch size \(=16\), embedding dimension \(=64\)), with early stopping based on validation PRAUC. For evaluation, we report results using bootstrapping over 10 random 80\% subsamples of the test set.
\paragraph{Self-RAG:}
Self-RAG~\cite{asai2024selfrag} integrates retrieval and generation through learned critique signals. We retrieve the top-7 similar patient visits via FAISS (all-MiniLM-L6-v2, inner-product similarity) and filter them using a trained ISREL classifier (LLaMA 3.1-8B + LoRA, $r=16$, $\alpha=32$), which is fine-tuned on $\sim$24K automatically labeled visit pairs using prescription ingredient overlap ($\geq0.3$ = Relevant) with DeepSpeed ZeRO-2 ($lr=2\times10^{-5}$, 3 epochs). At inference, each ISREL-passed candidate independently generates a prescription, scored by rule-based ISSUP (ingredient overlap with retrieved prescription) and ISUSE (prescription count appropriateness). The highest-scoring candidate ($\text{score} = 1.0 + \text{ISSUP} + 0.5 \times \text{ISUSE}$) is selected as the final prescription.
\paragraph{MultiQueryRetrieval:}
MultiQueryRetrieval~\cite{langchain_multiqueryretriever} is a training-free RAG baseline that improves retrieval recall by generating $n=3$ semantically diverse rewrites of each patient query, retrieving similar cases independently for each rewrite, and merging the results. Training admissions are indexed in a FAISS vector store using \texttt{all-MiniLM-L6-v2} embeddings; after deduplication across query variants, the top-$k=7$ cases above cosine similarity 0.9 are formatted as few-shot context and passed to the generator in a single forward pass.

\paragraph{Guideline RAG:}

\begin{table}[h]
\centering
\small
\caption{\textbf{Hyperparameters used in Guideline RAG experiments.}}

\label{tab:guidelinerag_hyperparameters}
\adjustbox{max width=\columnwidth}{
\begin{tabular}{lp{0.65\columnwidth}}
\toprule
\textbf{Parameter} & \textbf{Value} \\
\midrule
\multicolumn{2}{l}{\textit{Vector Database Settings}} \\
chunk size & 1200 \\ 
chunk overlap & 200 \\ 
Parkinson Guideline &~\cite{https://doi.org/10.1002/mds.27372, 10.1001/jama.2014.3654, https://doi.org/10.1002/mds.27602}  \\
MIMIC Guideline &~\cite{STEVENS2021e545, 10.1093/cid/ciw353, doi:10.1161/CIR.0000000000001063, 10.1164/rccm.201908-1581ST, Evans2021Surviving, 10.2337/dc26-SINT} \\
\midrule
\multicolumn{2}{l}{\textit{Retrieval Settings}} \\
Retrieval top-$k$  & 3  \\
Threshold $\tau$ & 0.3  \\
\bottomrule
\end{tabular}
}
\vspace{-0.3cm}
\end{table}

For Guideline RAG, we constructed a retrieval pool using external clinical practice guidelines and medical textbooks. 
Detailed hyperparameters for this process are documented in Table~\ref{tab:guidelinerag_hyperparameters}. 
We retrieve relevant guideline sections based on the patient's records and use them as supporting evidence.
For the Parkinson dataset, the retrieval query is constructed from the Subjective, Objective, and Assessment sections of the clinical note, whereas for the MIMIC dataset the query is derived from diagnosis-oriented visit text.

\paragraph{TreatRAG:}

Following the approach proposed in TreatRAG~\cite{treatrag_2025}, we implement a diagnosis-centered retrieval pipeline. 
For the Parkinson’s dataset, where clinical records are structured as SOAP notes, we prioritize the `Assessment' field for the query as it most closely mirrors the clinical diagnosis. In instances where the `Assessment' data is unavailable, the `Subjective' field is utilized as an alternative source for the query.
For the MIMIC-IV dataset, we utilize the \texttt{diagnoses} field from the current visit as the retrieval query.
To compute similarity, both the query and the candidate cases are first converted to lowercase and tokenized by whitespace. 
These tokens are then used to construct $N$-gram sets (specifically bigrams, where $N=2$).
The similarity between the query case ($A$) and a candidate case ($B$) is measured using the Jaccard similarity coefficient:
$$J(A,B)=\frac{|A\cap B|}{|A\cup B|}.$$
To mitigate noise from irrelevant matches, we apply adaptive filtering to the similarity scores. 
We retain only those cases with positive similarity that satisfy the threshold:$$J(A,B) \ge \mu + 1.5\sigma,$$where $\mu$ and $\sigma$ represent the mean and standard deviation of the similarity score distribution, respectively. 
The remaining candidates are then ranked by their scores to identify the most relevant historical cases for prescription generation.

\paragraph{MedReflect:}
While MedReflect~\cite{huang_2026_medreflect} was originally designed as a training framework to construct reflective supervision trajectories, we adapt its core logic into an inference-time refinement procedure. 
Instead of generating data for fine-tuning, we implement an inference-time refinement sequence that generates reflective questions regarding potential clinical errors and generates answers to these queries. 
These synthesized question-answer pairs are then used to refine the initial draft into a final prescription.
Detailed prompt templates are provided in Appendix~\ref{app:others_prompt}.

\subsection{Specialist Evaluation Protocol}
\label{app:human_eval_protocol}

To assess the clinical reasoning quality, we conducted a blinded pairwise comparison with three medical experts. 
For each case, experts evaluated anonymized responses from Model A and Model B based on the same patient record, selecting a preferred model or a tie. 
This evaluation comprised 30 question-answer pairs per comparison, with independent judgments collected across the following six clinical dimensions:

\begin{enumerate}
    \item \textbf{Clinical Accuracy:} Based on the patient's clinical note, which response provides the most appropriate medication recommendation for the patient?
    \item \textbf{Safety:} Which response is safer and less likely to cause harm to the patient?
    \item \textbf{Faithfulness:} Which response bases its reasoning on information actually present in the clinical note, without introducing unsupported facts?
    \item \textbf{Clinical Reasoning:} Which response demonstrates more appropriate clinical reasoning when selecting medications?
    \item \textbf{Clinical Usefulness:} Which response would be more helpful for supporting your clinical decision-making in practice?
    \item \textbf{Overall Preference:} Overall, which response would you prefer in clinical practice?
\end{enumerate}
Table~\ref{tab:survey_example} provides an example of input clinical note along with the reasoning outputs generated by PACE-RAG and the Guideline RAG. 
To ensure a fair comparison, we presented only the reasoning components (clinical evidence and the final prescription), omitting the patient summary and keywords from the evaluation materials.

\section{Additional Analysis}
\label{app:add_experiments}

\subsection{Balancing Precision and Recall}
\label{app:precision-recall-tradeoff-results}

\begin{table}[t]
\centering
\small
\setlength{\tabcolsep}{5.5pt}
\renewcommand{\arraystretch}{1.08}
\caption{
\textbf{Subgroup analyses of precision--recall trade-offs.}
We report performance by medication-history availability and prescription complexity.
}
\label{tab:subgroup_analysis}
\vspace{-2mm}

(a) Analysis by medication-history availability

\vspace{1mm}
\resizebox{\columnwidth}{!}{
\begin{tabular}{ll lccc}
\toprule
\textbf{Dataset} & \textbf{Subgroup} & \textbf{Method} 
& \textbf{Precision} & \textbf{Recall} & \textbf{Micro-F1} \\
\midrule

\multirow{6}{*}{MIMIC-IV}
& \multirow{3}{*}{History present}
& Zero-shot & 0.7202 & 0.2543 & 0.3398 \\
& & TreatRAG & \textbf{0.7328} & 0.2970 & 0.3846 \\
& & PACE-RAG & 0.6891 & \textbf{0.5182} & \textbf{0.5630} \\
\cmidrule(lr){2-6}
& \multirow{3}{*}{History absent}
& Zero-shot & 0.0943 & 0.0352 & 0.0510 \\
& & TreatRAG & \textbf{0.5183} & \textbf{0.1942} & \textbf{0.2472} \\
& & PACE-RAG & 0.1354 & 0.0371 & 0.0483 \\

\midrule

\multirow{6}{*}{HPH}
& \multirow{3}{*}{History present}
& Zero-shot & 0.5171 & 0.6613 & 0.5756 \\
& & TreatRAG & 0.4893 & \textbf{0.8979} & 0.6007 \\
& & PACE-RAG & \textbf{0.8353} & 0.8720 & \textbf{0.8399} \\
\cmidrule(lr){2-6}
& \multirow{3}{*}{History absent}
& Zero-shot & 0.0234 & 0.0365 & 0.0264 \\
& & TreatRAG & \textbf{0.2637} & \textbf{0.6003} & \textbf{0.3561} \\
& & PACE-RAG & 0.0234 & 0.0365 & 0.0267 \\

\bottomrule
\end{tabular}
}

\vspace{3mm}

(b) Analysis by prescription complexity

\vspace{1mm}
\resizebox{\columnwidth}{!}{
\begin{tabular}{ll lccc}
\toprule
\textbf{Dataset} & \textbf{Subgroup} & \textbf{Method} 
& \textbf{Precision} & \textbf{Recall} & \textbf{Micro-F1} \\
\midrule

\multirow{6}{*}{MIMIC-IV}
& \multirow{3}{*}{Simple ($\leq15$) }
& Zero-shot & 0.3941 & 0.1968 & 0.2655 \\
& & TreatRAG & \textbf{0.5691} & 0.3079 & 0.3885 \\
& & PACE-RAG & 0.4011 & \textbf{0.3480} & \textbf{0.4110} \\
\cmidrule(lr){2-6}
& \multirow{3}{*}{Complex ($>15$)}
& Zero-shot & 0.6112 & 0.1527 & 0.2314 \\
& & TreatRAG & \textbf{0.7536} & 0.2068 & 0.3127 \\
& & PACE-RAG & 0.5921 & \textbf{0.3442} & \textbf{0.4368} \\

\midrule

\multirow{6}{*}{HPH}
& \multirow{3}{*}{Simple ($\leq2$)}
& Zero-shot & 0.3955 & 0.6075 & 0.4627 \\
& & TreatRAG & 0.3995 & \textbf{0.8721} & 0.4923 \\
& & PACE-RAG & \textbf{0.7254} & 0.8038 & \textbf{0.7373} \\
\cmidrule(lr){2-6}
& \multirow{3}{*}{Complex ($>2$)}
& Zero-shot & 0.7422 & 0.6146 & 0.6791 \\
& & TreatRAG & 0.7055 & \textbf{0.8762} & 0.7720 \\
& & PACE-RAG & \textbf{0.9072} & 0.7973 & \textbf{0.8563} \\

\bottomrule
\end{tabular}
}

\vspace{-3mm}
\end{table}

To better understand the precision--recall trade-off, we conduct subgroup analyses based on medication-history availability and prescription complexity.
As shown in Table~\ref{tab:subgroup_analysis}~(a), PACE-RAG performs particularly well when recent medication history is available.
This is because PACE-RAG uses the patient's active medication history as a reference for determining whether to maintain, add, or remove medications, while using retrieved prescribing tendencies as supporting evidence for these decisions.
In contrast, TreatRAG directly incorporates prescription histories from retrieved similar patients, which can increase recall by predicting more medications but may also include medications that are less relevant to the current patient, as reflected in its lower precision.
When recent medication history is absent, PACE-RAG loses an important patient-specific reference for refinement and therefore tends to behave more conservatively, whereas TreatRAG can still increase recall by relying directly on prescriptions from retrieved cases.

Prescription complexity further shows that the precision--recall trade-off varies across datasets, but PACE-RAG maintains the strongest overall balance.
As shown in Table~\ref{tab:subgroup_analysis}~(b), PACE-RAG achieves the highest micro-F1 in both simple and complex cases across MIMIC-IV and HPH.
The improvement arises from different aspects depending on the dataset: PACE-RAG improves recall on complex MIMIC-IV cases, while achieving higher precision on complex HPH cases.
These results suggest that PACE-RAG improves the overall prescription-set prediction quality by balancing precision and recall, rather than by consistently favoring one side of the trade-off.

\subsection{Focus-specific retrieval maintains stable clinical evidence across retrieved cases}
\label{app:effect_of_retrieval}

\begin{figure}[t!]
    \centering
    \includegraphics[width=1\linewidth]{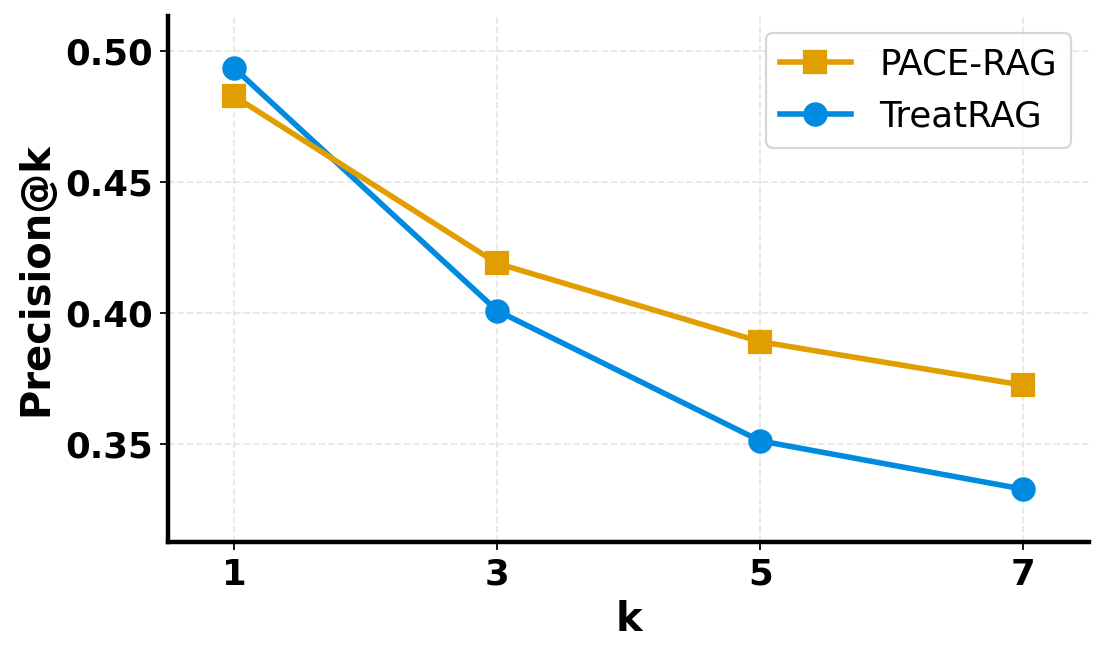}
    \caption{\textbf{Comparison of retrieval quality via drug precision@$k$.} 
    While TreatRAG accumulates irrelevant drugs as $k$ increases, 
    PACE-RAG maintains more stable precision by retrieving evidence based on patient-specific focus queries.
    }
    \label{fig:precision_k_graph}
    \vspace{-4mm}
\end{figure}

To assess retrieval quality, we employed the Drug Precision@$k$ metric (Figure~\ref{fig:precision_k_graph}), defined as the ratio of correctly matched medications to the total number of unique drugs extracted from the top-$k$ retrieved cases. 
TreatRAG relies on lexical overlap-based retrieval over similar patient records.
Thus, when the top-1 retrieved case happens to share many words or phrases with the target patient, TreatRAG can achieve high precision at $k=1$.
However, as $k$ increases, additional retrieved cases are more likely to include weakly matched or noisy examples.
Since TreatRAG incorporates the full prescription histories of similar patients, 
these additional cases introduce routine or chronic medications that are not directly relevant to the current clinical decision.
As a result, the retrieved evidence suffers from a \textit{majority bias}, where common background medications dominate the retrieval context, 
leading to a decline in Drug Precision as $k$ increases (from 49.3\% at $k=1$ to 33.2\% at $k=7$). 
In contrast, PACE-RAG retrieves evidence using patient-specific focus queries and semantic similarity, allowing it to focus on clinical signals relevant to the current patient rather than the entire medical history.
This approach effectively minimizes the accumulation of irrelevant chronic noise, maintaining a significantly higher precision baseline (37.2\% at $k=7$).
By providing precise pharmacological evidence, PACE-RAG enables the generator to produce more reliable clinical decisions, ultimately outperforming TreatRAG.

\subsection{Maintaining Performance with Simplified Prompts}
\label{app:robustness_prompt}

\begin{table}[t]
\centering
\small
\setlength{\tabcolsep}{8pt}
\renewcommand{\arraystretch}{1.1}
\caption{
\textbf{Robustness to simplified prompt variants.}
PACE-RAG maintains performance even when using simplified prompts for each stage.
}
\label{tab:prompt_robustness}
\vspace{-2mm}
\begin{tabular}{l|cc}
\toprule
\textbf{Prompt variant} & \textbf{F1 score} & \textbf{Acc} \\
\midrule
Simplified Stage 1 & 47.37\textcolor{gray}{$\pm$0.27} & 51.43\textcolor{gray}{$\pm$0.26}  \\
Simplified Stage 2 & 47.86\textcolor{gray}{$\pm$0.14} & 51.86\textcolor{gray}{$\pm$0.07}\\
Simplified Stage 3 & 43.76\textcolor{gray}{$\pm$0.34} & 48.54\textcolor{gray}{$\pm$0.37} \\
\bottomrule
\end{tabular}
\end{table}

Although PACE-RAG uses detailed prompts for each stage, its main contribution is not prompt engineering.
Instead, PACE-RAG contributes a patient-aware clinical RAG workflow that transforms retrieved patient cases into medication-level evidence and refines the final prescription through multiple stages.
To test whether the method depends on highly hand-crafted prompts, we evaluate simplified prompt variants for each stage on MIMIC-IV. 
The simplified prompts used in this experiment are provided in Figure~\ref{fig:simplified_prompts}.
As shown in Table~\ref{tab:prompt_robustness}, all simplified variants outperform the baseline methods and show performance close to the original PACE-RAG.
This suggests that the improvement mainly comes from the overall patient-aware retrieval and refinement workflow rather than from a specific prompt formulation.

\subsection{Reduction of Unsupported Medication Changes}
\label{app:medication_change_analysis}

\begin{table}[t]
\centering
\small
\setlength{\tabcolsep}{7pt}
\renewcommand{\arraystretch}{1.1}
\caption{
\textbf{Clinical support for medication changes.}
PACE-RAG produces fewer unsupported medication removals than TreatRAG while maintaining a comparable unsupported addition rate.
}
\label{tab:medication_change_analysis}
\vspace{-2mm}
\begin{tabular}{lcc}
\toprule
\textbf{Medication change type} 
& \textbf{TreatRAG} 
& \textbf{Ours} \\
\midrule
Unsupported additions ($\downarrow$) 
& 639 (4.3\%) 
& 631 (4.4\%) \\
Unsupported removals ($\downarrow$) 
& 4,821 (32.4\%) 
& 724 (5.1\%) \\
\bottomrule
\end{tabular}
\end{table}

To further assess clinical validity, we analyze whether each medication change made by the model is clinically supported.
Specifically, we use GPT-4o as an external judge to review each added or removed medication and determine whether the change is supported by the current clinical note, recent medication history, and retrieved patient evidence.
As shown in Table~\ref{tab:medication_change_analysis}, PACE-RAG produces a similar rate of unsupported additions compared with TreatRAG, but substantially fewer unsupported removals.
This suggests that PACE-RAG is less likely to make medication-change decisions without patient-specific clinical evidence.

\begin{table}[t]
\centering
\caption{
\textbf{Medication-level analysis of maintain, discontinue, and add decisions on the HPH and MIMIC-IV datasets.}
}
\label{tab:medication_decision_analysis}
\resizebox{\linewidth}{!}{
\begin{tabular}{ll|ccc}
\toprule
\textbf{Dataset} 
& \textbf{Metric(Acc.)}
& \textbf{Baseline}
& \textbf{TreatRAG}
& \textbf{PACE-RAG} \\
\midrule

\multirow{4}{*}{HPH}
& Decision
& 78.51
& 89.08
& \textbf{89.99} \\

& Maintain
& 85.16
& 98.61
& \textbf{99.85} \\

& Discontinue
& \textbf{18.75}
& 3.47
& 1.39 \\

& Add
& 5.43
& \textbf{15.84}
& 0.90 \\
\midrule

\multirow{4}{*}{MIMIC-IV}
& Decision
& 53.09
& 49.43
& \textbf{63.67} \\

& Maintain
& 49.45
& 43.08
& \textbf{93.40} \\

& Discontinue
& 59.92
& \textbf{61.38}
& 7.78 \\

& Add
& 0.43
& 1.04
& \textbf{1.54} \\

\bottomrule
\end{tabular}
}
\end{table}
\vspace{-5mm}

\begin{table}[t]
\centering
\caption{
\textbf{Evaluation on medical-specialized language models.}
We evaluate Baseline and PACE-RAG using MedGemma and HuatuoGPT.
}
\label{tab:medical_model}
\footnotesize
\setlength{\tabcolsep}{5pt}
\renewcommand{\arraystretch}{1.12}
\begin{tabular}{l|lcc}
\toprule
Model & Method & F1 & Acc \\
\midrule
MedGemma (4B) & Baseline & 69.99 & 74.21 \\
              & \textbf{PACE-RAG} & \textbf{77.43} & \textbf{79.58} \\
\midrule
HuatuoGPT (8B) & Baseline & 43.62 & 45.10 \\
               & \textbf{PACE-RAG} & \textbf{73.88} & \textbf{74.98} \\
\bottomrule
\end{tabular}
\end{table}

\subsection{Medication-Level Decision Analysis}
To further analyze medication-level decisions, we separately evaluate whether each method correctly maintains, discontinues, or adds medications. 
The results are reported in Table~\ref{tab:medication_decision_analysis} using the following four metrics:
\begin{itemize}
\item \textbf{Decision Accuracy}: accuracy of maintain-versus-discontinue decisions for medications in the active history.
\item \textbf{Maintain Accuracy}: proportion of medications correctly preserved.
\item \textbf{Discontinue Accuracy}: proportion of medications correctly removed.
\item \textbf{Add Accuracy}: proportion of newly prescribed medications correctly added.
\end{itemize}
PACE-RAG achieves the highest Decision Accuracy on both HPH and MIMIC-IV, reaching 89.99 and 63.67, respectively. 
Its performance differs across medication-change types: PACE-RAG shows strong accuracy for maintaining appropriate active medications and achieves the highest Add Accuracy on MIMIC-IV, whereas TreatRAG performs better on additions in HPH.
More broadly, the results show that addition and discontinuation performance varies substantially across methods and datasets, with no method consistently performing best across all medication-change types. 
These results indicate that medication-change prediction remains challenging and that the relative strengths of different methods depend on the type of treatment decision.
Overall, the action-level analysis complements the prescription-set metrics by providing a more detailed characterization of how PACE-RAG behaves across maintain, add, and discontinue decisions.

\subsection{Clinical Relevance of Extracted Medication Additions}
\label{subsec:clinical_relevance_medication}
To address concerns that Stage~2 might inadvertently extract unrelated background prescribing patterns, we conducted a relevance evaluation using an external GPT-4o-based clinical judge (prompts provided in Appendix~\ref{app:others_prompt}). 
The judge evaluated whether each extracted medication addition was clinically connected to the target clinical focus and appropriate for the patient's context.
We observed high relevance rates of 96.36\% on the HPH cohort and 83.52\% on the MIMIC-IV dataset. 
These results demonstrate that the medication additions extracted via Prescribing Tendency Analysis are clinically coherent with the target symptom or diagnosis, successfully filtering out unrelated baseline medications present in the retrieved cases.

\subsection{Effectiveness across specialized medical architectures}
\label{subsec:medical_model_performance}

Beyond general-purpose LLMs, we evaluate our framework on domain-specific models to verify its robustness in specialized clinical settings. As shown in Table~\ref{tab:medical_model}, we compare PACE-RAG against the baseline using MedGemma (4B)~\cite{sellergren2025medgemmatechnicalreport} and HuatuoGPT-o1 (8B)~\cite{chen2024huatuogpto1medicalcomplexreasoning}.
When tested on the HPH cohort, PACE-RAG yields substantial gains in both F1 score and accuracy across both architectures.
These results highlight that PACE-RAG’s effectiveness is model-agnostic; its ability to enhance clinical reasoning remains consistent regardless of whether the underlying architecture is a general-purpose or a specialized language model.

\section{Ablation Study}
\label{app:ablation}

\begin{table}[t]
\centering
\caption{
\textbf{Copy-History-based ablations evaluated using F1 score.}
Simple LLM adjustment or retrieval-based medication addition does not reproduce the performance of the structured PACE-RAG workflow.
}
\label{tab:copy_history_ablation}
\resizebox{\linewidth}{!}{
\begin{tabular}{lcc}
\toprule
\textbf{Setting} & \textbf{HPH F1} & \textbf{MIMIC-IV F1} \\
\midrule
Copy-History
    & 80.06
    & 40.57 \\

Copy-History + LLM adjustment
    & $78.00\pm1.20$
    & $34.62\pm0.13$ \\

Copy-History + retrieved additions
    & $77.41\pm0.24$
    & $39.86\pm0.08$ \\

Copy-History + focus-specific additions
    & $79.90\pm0.12$
    & $42.87\pm0.36$ \\

\midrule
\textbf{PACE-RAG}
    & $\textbf{80.84}\pm0.12$
    & $\textbf{47.22}\pm0.12$ \\
\bottomrule
\end{tabular}
}
\end{table}

\begin{table}[t]
\centering
\caption{
\textbf{Component-level ablations evaluated using F1 score.}
Each variant removes one design component from the complete PACE-RAG workflow.
}
\label{tab:component_ablation}
\resizebox{\linewidth}{!}{
\begin{tabular}{lcc}
\toprule
\textbf{Setting} & \textbf{HPH F1} & \textbf{MIMIC-IV F1} \\
\midrule
w/o focus-specific retrieval
    & $79.56\pm0.30$
    & $44.88\pm0.17$ \\

w/o prescribing tendency analysis
    & $79.17\pm0.35$
    & $40.86\pm0.17$ \\

w/o mandatory history preservation
    & $80.73\pm0.20$
    & $\textbf{47.46}\pm0.43$ \\

\midrule
\textbf{PACE-RAG}
    & $\textbf{80.84}\pm0.12$
    & $47.22\pm0.12$ \\
\bottomrule
\end{tabular}
}
\end{table}

\subsection{Copy-History-Based Ablation}
To examine whether PACE-RAG's performance can be explained primarily by conservative reuse of the patient's medication history, we construct three stronger variants of the \texttt{Copy-History baseline}.
(1) \texttt{Copy-History + LLM adjustment} revises the copied medication history using the current patient record. 
(2) \texttt{Copy-History + retrieved additions} augments the copied history with medications identified from globally retrieved patient cases, 
whereas (3) \texttt{Copy-History + focus-specific additions} uses medications obtained from focus-specific retrieved cases.
As shown in Table~\ref{tab:copy_history_ablation}, simply applying LLM-based refinement or appending medications from retrieved cases does not consistently improve upon Copy-History. 
Although focus-specific additions provide some benefit on MIMIC-IV, they still remain below the complete PACE-RAG workflow.
These results indicate that PACE-RAG's gains do not arise from medication-history reuse or retrieval alone. Rather, they depend on the structured integration of patient history, focus-specific retrieval, and prescribing tendency analysis.

\subsection{Component-Level Ablation}
We isolate the contributions of focus-specific retrieval, prescribing tendency analysis, and prompt of mandatory history preservation. 
As shown in Table~\ref{tab:component_ablation}, removing focus-specific retrieval or prescribing tendency analysis consistently reduces F1, with the largest degradation observed when prescribing tendency analysis is removed on MIMIC-IV.
In contrast, removing the mandatory history-preservation rule yields performance comparable to the full model. This indicates that PACE-RAG's performance is not simply driven by a prompt instructing the model to retain the previous medication history unchanged. 
Instead, its effectiveness depends on the structured combination of focus-specific retrieval and prescribing tendency analysis.

\begin{table}[t]
\centering
\caption{
\textbf{Perturbation analysis of focus extraction and case retrieval.}
Perturbing patient-specific clinical focuses or retrieved patient cases consistently reduces F1.
}
\label{tab:evidence_chain_perturbation}
\resizebox{\linewidth}{!}{
\begin{tabular}{lcc}
\toprule
\textbf{Setting}
& \textbf{HPH F1}
& \textbf{MIMIC-IV F1} \\
\midrule
Random focus
& $79.72 \pm 0.41$
& $45.07 \pm 0.29$ \\

Generic focus
& $79.18 \pm 0.33$
& $40.02 \pm 0.05$ \\
\midrule
Random retrieval
& $79.10 \pm 0.11$
& $44.79 \pm 0.30$ \\

Low-similarity retrieval
& $78.98 \pm 0.28$
& $46.74 \pm 0.08$ \\

\midrule
\textbf{PACE-RAG}
& $\mathbf{80.84 \pm 0.12}$
& $\mathbf{47.22 \pm 0.12}$ \\
\bottomrule
\end{tabular}
}
\end{table}

\begin{figure}[t]
    \centering
    \includegraphics[width=1\linewidth]{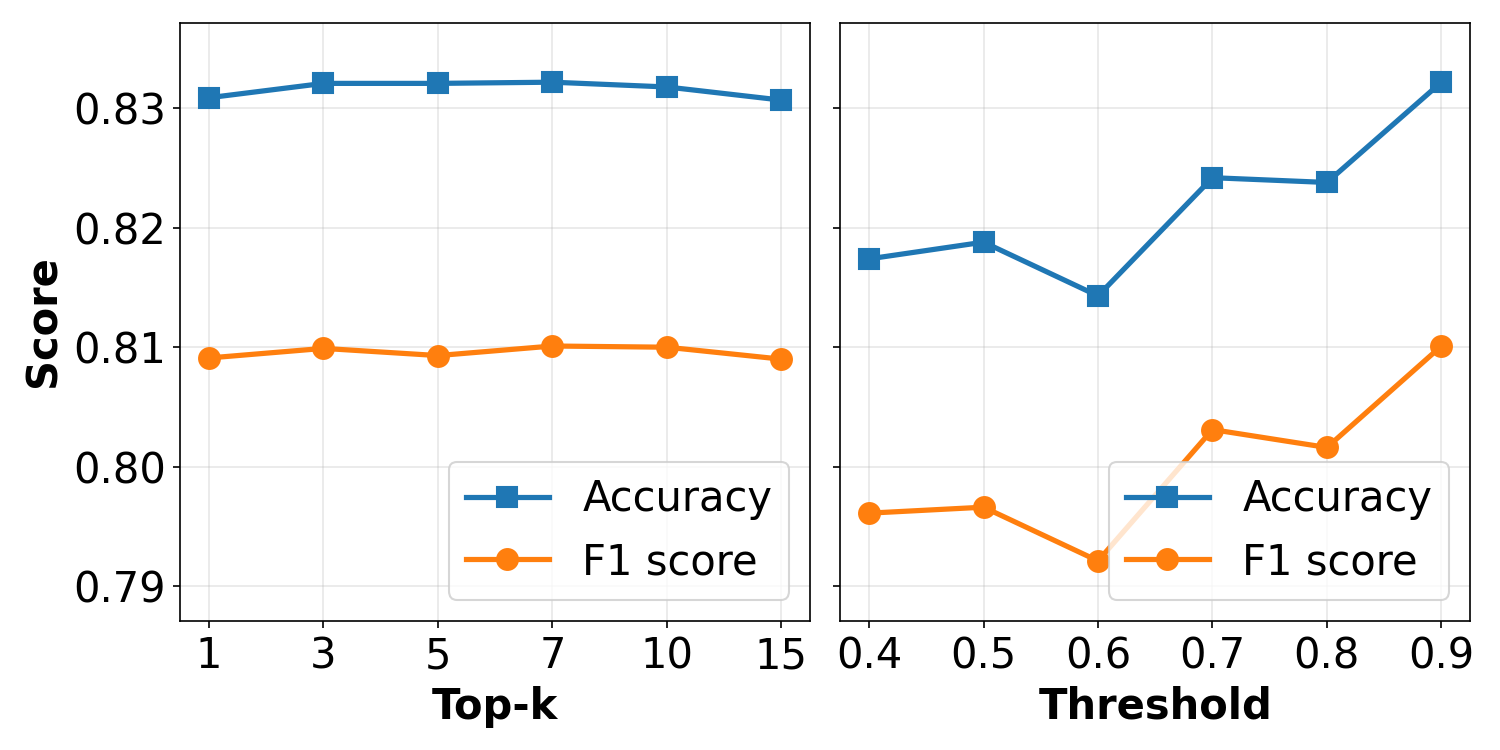}
    \caption{\textbf{Ablation of hyperparameters.} 
    Performance remains stable across various top-$k$ values (Left), while increasing the similarity threshold $\tau$ (Right) consistently enhances results.}
    \label{fig:performance_hyperparameter}
\end{figure}

\begin{table}[t]
\centering
\small
\setlength{\tabcolsep}{6pt}
\renewcommand{\arraystretch}{1.1}
\caption{
\textbf{Robustness analysis of retrieval hyperparameters.}
We evaluate the joint effect of the retrieval count $k$ and similarity threshold $\tau$, and further test robustness under noisy retrieval settings.
}
\label{tab:retrieval_hyperparameter_robustness}
\vspace{-2mm}

\textbf{(a) Joint effect of $k$ and $\tau$}

\vspace{1mm}
\begin{tabular}{c|ccc}
\toprule
\textbf{$\tau$} & \textbf{$k=3$} & \textbf{$k=7$} & \textbf{$k=10$} \\
\midrule
0.3 & 79.18 & 78.27 & 78.59 \\
0.6 & 79.59 & 79.21 & 78.59 \\
0.9 & 80.99 & 81.01 & 81.00 \\
\bottomrule
\end{tabular}

\vspace{3mm}

\textbf{(b) Noisy retrieval settings}

\vspace{1mm}
\begin{tabular}{cc|cc}
\toprule
\textbf{$k$} & \textbf{$\tau$} & \textbf{Sim. Patient RAG} & \textbf{Ours} \\
\midrule
10 & 0.3 & 76.55 & 78.59 \\
7  & 0.3 & 76.14 & 78.27 \\
\bottomrule
\end{tabular}

\end{table}

\begin{table}[t]
\centering
\small
\setlength{\tabcolsep}{8pt}
\renewcommand{\arraystretch}{1.1}
\caption{
\textbf{Sensitivity to the maximum number of extracted focus queries.}
PACE-RAG remains stable across different limits on the number of extracted focus queries.
}
\label{tab:max_focus_query_sensitivity}
\vspace{-2mm}
\resizebox{\columnwidth}{!}{
\begin{tabular}{c|cccc}
\toprule
\textbf{Max \# queries} & \textbf{F1 score}  & \textbf{Acc}  & \textbf{Precision}  & \textbf{Recall} \\
\midrule
1 & 80.70  & 82.90 & 83.84 & 81.06 \\
2 & 81.01 & 83.22 & 84.07 & 81.54 \\
3 & 80.90  & 82.97  & 81.31 & 84.00 \\
5 & 80.85  & 82.87  &  83.83 &  81.48 \\
\bottomrule
\end{tabular}
}
\end{table}

\subsection{Evidence-Chain Perturbation Ablation.}
We conduct perturbation tests to assess whether patient-specific clinical focuses and relevant retrieved cases contribute meaningful evidence. 
We replace the extracted focus with either a random focus from another patient or a generic diagnosis, such as ``Parkinson's disease.'' We also replace the retrieved cases with either randomly selected or low-similarity patients.
As shown in Table~\ref{tab:evidence_chain_perturbation}, degrading either the clinical focus or the retrieved evidence generally reduces F1 on both datasets. The larger drops under the generic-focus and random-retrieval settings indicate that using patient-specific clinical focuses and clinically relevant similar-patient cases is important for prescription refinement.
These results show that PACE-RAG benefits not merely from adding retrieval context, but from retrieving evidence centered on the target patient's clinical features.

\subsection{Hyperparameter Sensitivity}
\label{app:hyperparamter_sensitivity}
We conduct all hyperparameter sensitivity analyses on the HPH dataset. First,
we evaluated the impact of the retrieval count ($k$) and the similarity threshold ($\tau$). 
As shown in Figure~\ref{fig:performance_hyperparameter}, performance remains notably stable across various $k$ values, suggesting that the framework is robust to the number of retrieved cases. 
In contrast, increasing $\tau$ consistently improves performance by filtering out clinically irrelevant noise. 
We further conduct a joint analysis of $k$ and $\tau$ in Table~\ref{tab:retrieval_hyperparameter_robustness}~(a).
PACE-RAG remains stable across different retrieval counts, with F1 scores ranging from 78.27 to 81.01, and consistently performs best under the stricter threshold $\tau=0.9$.

Because lower similarity thresholds and larger retrieval counts may introduce less relevant patient cases, we further test PACE-RAG under noisy retrieval settings.
As shown in Table~\ref{tab:retrieval_hyperparameter_robustness}~(b), PACE-RAG consistently outperforms Similar-Patient RAG even under noisy retrieval settings with a low $\tau$ and larger $k$, demonstrating its robustness to noisy retrieved contexts.

Second, we vary the maximum number of extracted focus queries in Table~\ref{tab:max_focus_query_sensitivity}.
PACE-RAG shows stable performance across different number of queries.
This is because the Stage 1 extractor is designed to select only clinically actionable focus queries for the current patient; therefore, increasing the maximum query limit does not substantially change the retrieved evidence unless additional meaningful clinical focus points are present.

\subsection{Sparse versus Dense Retrieval}
\label{app:sparse_dense_retrieval}

\begin{table}[t]
\small
\centering
\caption{\textbf{Ablation of retrieval strategies.} Keyword sparse retrieval vs.\ vector dense retrieval using HPH.}
\label{tab:method_retrieval}

\resizebox{\columnwidth}{!}{
\begin{tabular}{l|cccc}
\toprule
\textbf{Method}           & \textbf{F1 score} & \textbf{Acc} & \textbf{Precision} & \textbf{Recall} \\
\midrule
Sparse Retrieval & 0.7904 & 0.8137 & 0.8166 & 0.8022 \\
Dense Retrieval  & 0.8101 & 0.8322 & 0.8407 & 0.8154 \\
\bottomrule
\end{tabular}
}

\vspace{-6mm}
\end{table}

We compared our setting dense retrieval (vector embeddings) against a sparse retrieval (BM25). 
As summarized in Table~\ref{tab:method_retrieval}, dense retrieval consistently outperforms sparse retrieval across all metrics. These results confirm that the semantic-aware embeddings provide more relevant and contextually rich patient evidence than literal keyword matching.

\section{Prompt Templates}
\label{app:prompts}


\subsection{Initial Draft Generation \& Zero-Shot Baseline Prompts}
\label{app:prompts_initial_draft}

The prompts in Figure~\ref{fig:prompt_initial_draft} serve a dual purpose: they act as the standalone `Zero-Shot Baseline' and function as the `Initial Prescription Generator' in PACE-RAG. 
\subsection{Stage 1 : Focus Query Extractor}
\label{app:prompts_focus_query_extractor}
Figure~\ref{fig:full_prompt_extractor} presents the Stage 1 prompts. 
The Focus Query Extractor extracts actionable clinical keywords (e.g., severe symptoms) while filtering out noise. 
Crucially, Extractor returns an empty list for stable patients, ensuring dynamic RAG is triggered only when medication adjustments are needed.

\subsection{Stage 2: Prescribing Tendency Analysis}
\label{app:prompts_tendency_analysis}
Figure~\ref{fig:full_prompt_analyzer} details the Stage 2 prompts. 
This stage strictly extracts \textit{treatment tendencies}—newly added medications around similar patients for the target focus. 
To isolate precise interventions, the prompt applies a strict causality check and filters out pre-existing medications. Furthermore, the model is strictly instructed to return an empty output unless a new medication is explicitly added for the target symptom, effectively preventing hallucinations when clinical evidence is insufficient.

\subsection{
Stage 3: Evidence-Constrained Prescription Refinement}
\label{app:prompt_stage3}
Figure~\ref{fig:full_prompt_verifier} outlines the Stage 3 prompts. 
This stage uses a rigid two-step mechanical algorithm: 
Step 1 preserves the patient's active history by forcing all currently taken medications to be retained in the final prescription.
Step 2 evaluates newly proposed medications, which represent medications frequently introduced in similar patient cases retrieved from the database.

\subsection{Stage 4: Explainable Clinical Summary}
\label{app:doctor_summary}
Figure~\ref{fig:full_prompt_summary} shows the Stage 4 prompt. The LLM synthesizes the patient's state, retrieved tendencies, and the verification log into a summary report.

\subsection{Others}
\label{app:others_prompt}

\paragraph{Keyword Extract Evaluation.}

We utilized GPT-4o in an LLM-as-a-judge capacity to assess the Stage 1 Focus Query Extractor, setting the temperature to 0.0 for consistency. 
As shown in Figure~\ref{fig:prompt_keyword_evaluator}, the judge model assigns a score from 1 to 5 based on how accurately the extractions align with the ground-truth prescription.

\paragraph{Baseline Methods.}

Figures~\ref{fig:prompt_treatrag_baseline} and~\ref{fig:prompt_guideline_baseline} present the prompts for the TreatRAG and GuidelineRAG baselines, respectively. 
Additionally, Figure~\ref{fig:prompt_medreflect} details the prompt for MedReflect, a self-correction baseline that employs an internal three-stage reflection pipeline consisting of Questioning, Answering, and Refining.

\paragraph{Simplified Prompts.}
We provide the simplified prompts used in the robustness experiment in Appendix~\ref{app:robustness_prompt}, with the corresponding prompt templates shown in Figure~\ref{fig:simplified_prompts}.
These simplified prompts correspond to the Stage 1 Focus Query Extractor, Stage 2 Prescribing Tendency Analysis, and Stage 3 Evidence-Constrained Prescription Refinement prompt for the MIMIC-IV dataset. Complex reasoning constraints and extraction rules were removed to evaluate the baseline stability of the pipeline.

\paragraph{Prescribing Tendency Relevance Evaluation.}
We used GPT-4o as an external LLM-as-a-judge (temperature~$0.0$) to assess the clinical coherence of medications marked as \texttt{ADD} in Stage~2 Prescribing Tendency Analysis.
For each unique $(\textit{focus}, \textit{medication})$ pair, the judge determined whether the addition was clinically connected to the retrieval focus and/or the patient's co-occurring context under an inclusive clinical rubric.
As shown in Figure~\ref{fig:prompt_medication_relevance_judge}, the judge returns a binary relevance decision with a short rationale; when available, a short patient-context snippet is included in the judgment input.

\begin{table*}[p]
\centering
\caption{\textbf{List and count of unique medications by dataset.}}
\label{tab:medications_list}
\adjustbox{max width=\textwidth}{
\begin{tabular}{lp{12cm}c}
\toprule
\textbf{Dataset Split} & \textbf{Medication Names} & \textbf{Count} \\
\midrule
\textbf{Parkinson Retrieval Pool} & 
Agomelatine, Amantadine sulfate, Aripiprazole, Benserazide HCl, Benztropine mesylate, Bupropion HCl, Carbidopa, Carbidopa monohydrate, Clozapine, Donepezil, Entacapone, Escitalopram, Ginkgo biloba ext, Levodopa, Memantine HCl, Opicapone, Paroxetine, Pramipexole, Procyclidine HCl, Quetiapine, Rasagiline, Risperidone, Rivastigmine, Ropinirole, Safinamide mesilate, Sertraline HCl, Tianeptine sodium, Trazodone HCl, Trihexyphenidyl HCl, Vortioxetine, carbidopa 
& 31 \\
\addlinespace
\textbf{Parkinson Test Set} & 
Amantadine sulfate, Aripiprazole, Benserazide HCl, Bupropion HCl, Carbidopa, Carbidopa monohydrate, Donepezil, Entacapone, Escitalopram, Levodopa, Memantine HCl, Olanzapine, Opicapone, Pramipexole, Procyclidine HCl, Quetiapine, Rasagiline, Rivastigmine, Ropinirole, Safinamide mesilate, Tianeptine sodium, Trazodone HCl, Trihexyphenidyl HCl, carbidopa 
& 24 \\
\midrule
\textbf{MIMIC Retrieval Pool} & 
3-oxoandrosten (4) derivatives, ACE inhibitors, plain, Acetic acid derivatives and related substances, Acid preparations, Actinomycines, Adamantane derivatives, Adrenergic and dopaminergic agents, Adrenergics in combination with corticosteroids or other drugs, excl. anticholinergics, Agents for dermatitis, excluding corticosteroids, Aldehydes and derivatives, Aldosterone antagonists, Alkyl sulfonates, Alpha and beta blocking agents, Alpha glucosidase inhibitors, Alpha-adrenoreceptor antagonists, Aluminium compounds, Amides, Amino acids, Amino acids and derivatives, Aminoalkyl ethers, Aminoquinolines, Aminosalicylic acid and derivatives, Aminosalicylic acid and similar agents, Androstan derivatives, Angiotensin II receptor blockers (ARBs), other combinations, Angiotensin II receptor blockers (ARBs), plain, Anilides, Anthracyclines and related substances, Anti-androgens, Anti-estrogens, Antiallergic agents, excl. corticosteroids, \dots 
& 406 \\
\addlinespace
\textbf{MIMIC Test Set} & 
ACE inhibitors, plain, Acetic acid derivatives and related substances, Acid preparations, Adrenergic and dopaminergic agents, Adrenergics in combination with corticosteroids or other drugs, excl. anticholinergics, Agents for dermatitis, excluding corticosteroids, Aldosterone antagonists, Alpha and beta blocking agents, Alpha-adrenoreceptor antagonists, Aluminium compounds, Amino acids, Amino acids and derivatives, Aminoquinolines, Aminosalicylic acid and similar agents, Angiotensin II receptor blockers (ARBs), other combinations, Angiotensin II receptor blockers (ARBs), plain, Anilides, Anthracyclines and related substances, Anti-estrogens, Antiallergic agents, excl. corticosteroids, Antiarrhythmics, class III, Antiarrhythmics, class Ib, Antiarrhythmics, class Ic, Antibiotics, Anticholinergics, Anticholinesterases, Antidotes, Antihidrotics, Antihistamines for topical use, Antiinfectives and antiseptics for local oral treatment, Antiinfectives for treatment of acne, \dots 
& 254 \\
\bottomrule
\end{tabular}
}
\end{table*}

\begin{table*}[p]
\centering
\small
\renewcommand{\arraystretch}{1.5} 
\caption{\textbf{Example of a survey questionnaire item for clinical evaluation.} Real-world patient cases were presented to clinical experts along with blinded outputs from Guideline RAG and PACE-RAG.}
\label{tab:survey_example}
\vspace{-3mm}
\begin{tabularx}{\textwidth}{l p{2.5cm} X}
\toprule
\multicolumn{3}{@{}p{\linewidth}@{}}{
\textbf{[Input Patient Case]}\newline
\textbf{Current Visit:}\newline
Subjective: External hypertension medication, Nomaloc and aspirin 100mg (external medication). My leg still feels numb.\newline
Objective: None\newline
Assessment: \#1. Right MCA infarction  \#2. HTN \#3. Parkinsonism \#4. Thrombocytopenia\newline
\newline
\textbf{Past Visit History (Most recent 3 visits):}\newline
7th Visit)\newline
Subjective: The right leg keeps pulling. The tremors are getting worse.\newline
Objective: None\newline
Assessment: Parkinson's disease, atypical cognitive IPD differential diagnosis is not possible (MSA possibility)\newline
Plan: Amantadine sulfate\newline
8th Visit)\newline
Subjective: has diarrhea.  It's hard to keep balance on legs.\newline
Objective: tandem gait : intact Whole body pain is present.\newline
Assessment: Parkinsonsim  Atypical cognitive IPD cognitive differentiation is not possible (MSA possible)\newline
Plan: Amantadine sulfate\newline
9th Visit)\newline
Subjective: when doing difficult tasks, hands shake a lot.\newline
Objective: tandem gait; impairment of bradykinesia: equivocal\newline
Assessment: Parkinson's disease, atypical, IPD, cognitive differentiation is not possible (MSA is possible) - early stage H/O right MCA territory infarction\newline
Plan: Amantadine sulfate
} \\
\midrule
\textbf{Method} & \textbf{Prescription} & \textbf{Reasoning} \\
\midrule

Guideline RAG 
& Amantadine sulfate
& * Patient summary *\newline
The patient has a history of external hypertension medication, Nomaloc and aspirin 100mg, and continues to experience numbness in the leg. They have a history of right MCA infarction, hypertension, parkinsonism, and thrombocytopenia. The patient is currently stable with no acute changes in symptoms. \newline
* Clinical Evidence * \newline
Amantadine sulfate is standard for managing Parkinsonism symptoms, particularly in patients with concurrent stroke (Right MCA infarction). Evidence supports its use in reducing bradykinesia and rigidity, though its efficacy for leg numbness is less defined. Aspirin and Nifedipine are consistent with secondary prevention for HTN and cerebrovascular risk. \newline
* Prescribe * \newline
Amantadine sulfate : Continued from patient's history for Parkinsonism management. Clinical evidence supports its use in reducing Parkinsonian motor symptoms, though its role in leg numbness remains unclear. No adjustment is needed given stable symptoms and absence of worsening neurological deficits. \\

\midrule

\textbf{PACE-RAG (Ours)} 
&Amantadine sulfate\newline
& * Patient summary *\newline
The patient has a history of external hypertension medication, Nomaloc and aspirin 100mg, and continues to experience numbness in the leg. They have a known right MCA infarction, hypertension, parkinsonism, and thrombocytopenia. The patient is currently stable with no acute changes in symptoms or neurological status.\newline
* Key word *\newline
Right MCA infarction, HTN, Parkinsonism, Thrombocytopenia, Leg numbness\newline
* Clinical Evidence * \newline
In similar cases, Amantadine sulfate has been used for managing symptoms of parkinsonism, particularly for dyskinesia and off-period symptoms. It is also used in conjunction with other medications like Levodopa/Carbidopa and Trihexyphenidyl for comprehensive symptom management. The drug is generally well-tolerated, though long-term use may lead to dyskinesia. \newline
* Prescribe * \newline
Amantadine sulfate : Amantadine sulfate is being continued from the patient's history as it is effective for managing symptoms of parkinsonism, particularly for dyskinesia and off-period symptoms. The decision to continue the medication is based on the patient's stable condition and the absence of acute changes in symptoms. Patterns from similar cases indicate that Amantadine sulfate is often used as an adjunct in managing parkinsonism, especially when combined with other medications like Levodopa/Carbidopa. The recommendation is to maintain the current regimen as the patient is stable and shows no signs of deterioration. \\

\bottomrule
\end{tabularx}
\vspace{-5mm}
\end{table*}


\begin{figure*}[p]
    \centering
    \begin{tcolorbox}[colback=parkinson!8!white, colframe=parkinson!80!black, title={Initial Prescription Generator / Zero-Shot Baseline: Parkinson's Disease}, fonttitle=\bfseries]
    \scriptsize
    \textbf{System Prompt:} \\
    You are a Parkinson's disease medication specialist. \\
    Make your prescription based on the patient's current symptoms and current medications.
    If there is active history, start from that active history! Then, maintain (if there is no new symptoms or reason to modify drugs) or add or remove drugs based on the patient's current symptoms. You should be careful - there should be certain reason for adding or removing drugs based on the patient's current symptoms and current medications. \\
    
    You should put one drug name at every line! \\
    \textbf{FORMAT STRICT:} Output ONLY the [START]...[END] block. No extra text. \\
    
    \textbf{STABILITY RULE (no active history):}
    \begin{itemize}[leftmargin=*, noitemsep, topsep=0pt]
        \item If symptoms are stable/improved/no major worsening, output only conservative drug.
    \end{itemize}
    
    \textbf{Output Format:} \\
    \texttt{[START]} \\
    \texttt{(Drug Name) | (short reason in 10 words or less)} \\
    \texttt{[END]} \\
    
    \textbf{User Prompt:} \\
    \textbf{Clinical Note:} Subjective: \{subjective\}, Objective: \{objective\}, Assessment: \{assessment\} \\
    \textbf{Most Recent Medications:} \{history\} \\
    
    \textbf{Past Clinical Visits (up to 3 visits before current):} \{recent\_visit\_history\_text\} \\
    
    \textbf{Task:} Generate prescription. Should put at least one drug in output.
    \end{tcolorbox}

    \vspace{-4mm}

    \begin{tcolorbox}[colback=mimic!8!white, colframe=mimic!90!black, title={Initial Prescription Generator / Zero-Shot Baseline: MIMIC-IV}, fonttitle=\bfseries]
    \scriptsize
    \textbf{System Prompt:} \\
    You are a clinical medication specialist for complex hospital inpatients (MIMIC-IV dataset). \\
    
    Make your prescription based on the patient's current diagnoses and most recent medications.
    If there is a most recent medications list, start from that list.
    Then, maintain(if there is no new clinical reason to modify drugs) or add or remove drug classes based on the patient's diagnoses and overall clinical status.
    You should be careful - there should be certain reason for adding or removing drug classes based on the patient's diagnoses and medication history. \\
    
    Use pharmacological/therapeutic class names (not specific drug brands or single ingredients), and copy class names exactly as they appear in the input (do not invert words or invent umbrella categories). \\
    
    You should put one drug class name at every line! \\
    \textbf{FORMAT STRICT:} Output ONLY the [START]...[END] block. No extra text. \\
    
    \textbf{STABILITY RULE (no most recent medications list):}
    \begin{itemize}[leftmargin=*, noitemsep, topsep=0pt]
        \item If the clinical picture is stable/improved/no major new problems, output only a conservative minimal set of classes.
        \item Prefer continuing existing chronic medication classes when they are clearly indicated.
        \item Avoid starting broad new classes unless there is clear diagnostic support.
    \end{itemize}
    
    \textbf{OUTPUT:} \\
    \texttt{[START]} \\
    \texttt{(Drug Class Name) | (short reason in 10 words or less)} \\
    \texttt{(Drug Class Name) | (short reason in 10 words or less)} \\
    \texttt{[END]}
    
    \tcblower
    \scriptsize
    
    \textbf{User Prompt:} \\
    \textbf{Patient Information:} \\
    - Diagnoses: \{diagnoses\} \\
    - Most Recent Medications: \{medications\} \\
    
    \textbf{Past Clinical Visits (up to 3 visits before current):} \\
    \{recent\_visit\_history\_text\} \\
    
    \textbf{Task:} Generate prescription. Should put at least one drug in output.
\end{tcolorbox}
\vspace{-4mm}
    \caption{\textbf{Prompt designs for the Initial Prescription Generator}, which also serve as the standalone Zero-Shot Baseline (No RAG). The prompts impose strict output formats, ontology mapping constraints (for MIMIC-IV), and conservative clinical heuristics.}
    \label{fig:prompt_initial_draft}
\end{figure*}


\begin{figure*}[p]
    \centering
    \begin{tcolorbox}[colback=parkinson!8!white, colframe=parkinson!80!black, title={Focus Query Extractor: Parkinson's Disease}, fonttitle=\bfseries]
    \scriptsize
    \textbf{System Prompt:} \\
    You are a precise Medical Symptom Extractor. Your ONLY job is to identify \textbf{NEW, WORSENING, or UNRESOLVED} pathological symptoms that require a medication change. \\
    You are NOT a doctor. Do NOT invent, infer, or hallucinate. \\

    \textbf{STRICT EXTRACTION RULES:}
    \begin{enumerate}[leftmargin=*, noitemsep, topsep=0pt]
        \item \textbf{THE `EMPTY' RULE (CRITICAL):}
        \begin{itemize}[noitemsep]
            \item If the patient is STABLE, IMPROVING, or ``feeling better''...
            \item If the text is just conversational (e.g., ``test results explained'', ``stopped drinking'')...
            \item $\rightarrow$ YOU MUST OUTPUT AN EMPTY LIST: \texttt{\{"keywords": []\}}
        \end{itemize}
        \item \textbf{NO BROAD DIAGNOSES:} NEVER output broad diseases like ``Parkinson's disease'' or ``Vascular Parkinsonism'' alone. Broad keywords ruin search databases. Only extract if attached to a specific severe symptom (e.g., ``Parkinson's disease with severe resting tremor'').
        \item \textbf{LITERAL MATCH:} Use exact words from the text.
        \item \textbf{LIMIT:} Max 2 phrases. Focus ONLY on acute issues.
    \end{enumerate}

    \textbf{Output format:} \\
    \texttt{\{"keywords": ["Target Symptom 1"]\}} OR \texttt{\{"keywords": []\}}

    \tcblower
    \scriptsize

    \textbf{User Prompt:} \\
    \textbf{Current Patient Input (Subjective/Objective/Assessment):} \\
    \{text\} \\
    
    \textbf{Active History:} \\
    \{active\_history\} \\
    
    Analyze the CURRENT text. If the patient is stable/improving/no acute symptoms, return \texttt{\{"keywords": []\}}. Otherwise, extract MAX 2 severe symptom phrases. Return JSON only.
    \end{tcolorbox}

    \vspace{-0.4cm}

    \begin{tcolorbox}[colback=mimic!8!white, colframe=mimic!90!black, title={Focus Query Extractor: MIMIC-IV}, fonttitle=\bfseries]
    \scriptsize
    \textbf{System Prompt:} \\
    You are a precise Medical Diagnosis Extractor for MIMIC-IV inpatients. Your ONLY job is to identify CURRENT ACTIVE diagnoses that are NEW, WORSENING, or UNRESOLVED and clearly matter for treatment. \\
    You are NOT a doctor. Do NOT invent, infer, or hallucinate. \\

    \textbf{STRICT EXTRACTION RULES:}
    \begin{enumerate}[leftmargin=*, noitemsep, topsep=0pt]
        \item \textbf{THE `EMPTY' RULE (CRITICAL):}
        \begin{itemize}[noitemsep]
            \item If the diagnoses list is empty, purely administrative/meta, or only screening/history/status phrases (e.g., ``Inpatient'', ``Follow-up visit'', ``History of X'')...
            \item If nothing clearly reflects an acute or active condition needing management...
            \item $\rightarrow$ YOU MUST OUTPUT AN EMPTY LIST: \texttt{\{"keywords": []\}}
        \end{itemize}
        \item \textbf{NO GENERIC LABELS OR FRAGMENTS:}
        \begin{itemize}[noitemsep]
            \item Do NOT output encounter/status/meta phrases alone (e.g., ``Inpatient'', ``Hospitalization'', ``Follow-up visit'').
            \item Forbidden single bare words: ``acute'', ``chronic'', ``failure'', ``infection'' by themselves.
        \end{itemize}
        \item \textbf{LITERAL MATCH:}
        \begin{itemize}[noitemsep]
            \item Always keep full diagnosis phrases exactly as written (e.g., ``Acute kidney failure'', ``Sepsis due to pneumonia'').
            \item Do NOT chop off qualifiers like ``acute'', ``chronic'', ``unspecified''.
            \item Every keyword must be an exact substring from the diagnoses text (case-insensitive). No paraphrasing or rewording.
        \end{itemize}
        \item \textbf{LIMIT:} Max 2 phrases. Focus ONLY on the most critical acute/active issues that drive treatment decisions.
    \end{enumerate}

    \textbf{OUTPUT FORMAT (JSON ONLY):} \\
    \texttt{\{"keywords": ["Diagnosis phrase 1", "Diagnosis phrase 2"]\}} OR \texttt{\{"keywords": []\}} \\
    No explanations, no markdown, no extra keys.

    \tcblower
    \scriptsize

    \textbf{User Prompt:} \\
    Diagnoses list: \\
    \{text\} \\

    \textbf{Active History:} \\
    \{active\_history\} \\
    
    Analyze the CURRENT text. If the patient is stable/improving/no acute symptoms, return \texttt{\{"keywords": []\}}. Otherwise, extract MAX 2 severe symptom phrases. Return JSON only.
    \end{tcolorbox}
    \vspace{-0.4cm}
    \caption{\textbf{Full prompts for Focus Query Extraction (Stage 1). }
    The extractor is designed to isolate the most critical, actionable clinical signals---such as severe symptoms and active pathologies---while filtering out broad diagnoses and administrative noise. 
    Furthermore, the model generates an empty list (\texttt{[]}) when patients are stable or improving. 
    This dual mechanism ensures dynamic RAG triggering, selectively engaging the retrieval process only when therapeutic adjustments are genuinely required.}
    \label{fig:full_prompt_extractor}
\end{figure*}


\begin{figure*}[p]
    \centering
    \begin{tcolorbox}[colback=parkinson!8!white, colframe=parkinson!80!black, title={Prescribing Tendency Analysis: Parkinson's Disease}, fonttitle=\bfseries]
    \scriptsize
    \textbf{System Prompt:} \\
    You are a highly focused Clinical Pattern Analyzer for Parkinson's Disease. \\
    
    \textbf{OBJECTIVE:} Analyze similar patient cases to identify EXACTLY which medications physicians \textit{NEWLY PRESCRIBED (ADDED)} to resolve the specific target symptom. \\
    
    \textbf{STRICT ANALYSIS RULES:}
    \begin{enumerate}[leftmargin=*, noitemsep, topsep=0pt]
        \item \textbf{FOCUS ONLY ON ADDITIONS:} Your ONLY job is to find drugs that were \textit{newly prescribed} (added) in the similar cases to treat the specific Target Clinical Focus.
        \item \textbf{CAUSALITY CHECK (CRITICAL):} The drug MUST have been added specifically for the Target Symptom. If the target is ``Falls'', do not extract antidepressants given for other reasons.
        \item \textbf{IGNORE MAINTAINED DRUGS:} Do not extract drugs that the patient was already taking.
        \item \textbf{CAUTIOUS EMPTY DEFAULT:} If the cases do not explicitly show a new drug being added to treat the target symptom, your \texttt{common\_additions} MUST be empty \texttt{[]}.
        \item \textbf{No Invention:} Rely ONLY on the provided text.
    \end{enumerate}
    
    \textbf{OUTPUT FORMAT:} Output ONLY valid JSON. \\
    \texttt{\{"dominant\_pattern": "ADD", "common\_additions": ["Drug A"], "reasoning": "Brief 1-sentence reason based on cases."\}}

    \tcblower
    \scriptsize
    
    \textbf{User Prompt:} \\
    \textbf{Current Patient Symptoms:} \{symptoms\} \\
================================================== \\
TARGET CLINICAL DIAGNOSIS: $>>>$ \{focus\_txt\} $<<<$ \\
* IMPORTANT: Filter the 'Similar Patient Cases' below. Only pay attention to how physicians treated THIS EXACT TARGET. \\
* Ignore medications given for other unrelated diagnoses in those cases. \\
================================================== \\
\textbf{Similar Patient Cases:} \{rag\_cases\} \\
    
    Analyze explicitly what was ADDED. Output JSON only.
    \end{tcolorbox}

    \vspace{-0.4cm}

    \begin{tcolorbox}[colback=mimic!8!white, colframe=mimic!90!black, title={Prescribing Tendency Analysis: MIMIC-IV}, fonttitle=\bfseries]
    \scriptsize
    \textbf{System Prompt:} \\
    You are a highly focused Clinical Pattern Analyzer for complex hospital inpatients (MIMIC-IV). \\
    
    \textbf{OBJECTIVE:} Analyze similar patient cases to identify EXACTLY which medication CLASSES physicians \textit{NEWLY PRESCRIBED (ADDED)} to resolve the specific target diagnosis. \\
    
    \textbf{STRICT ANALYSIS RULES:}
    \begin{enumerate}[leftmargin=*, noitemsep, topsep=0pt]
        \item \textbf{FOCUS ONLY ON ADDITIONS:} Your ONLY job is to find drug classes that were \textit{newly prescribed} (added) in the similar cases to treat the specific Target Clinical Focus.
        \item \textbf{CAUSALITY CHECK (CRITICAL):} The drug class MUST have been added specifically for the Target Diagnosis. (Note: General inpatient care classes like `Heparin group' or `laxatives' that consistently accompany the diagnosis across many similar cases as standard co-management or prophylaxis should also be extracted.)
        \item \textbf{IGNORE MAINTAINED DRUGS:} Do not extract drug classes that the patient was already taking.
        \item \textbf{CAUTIOUS EMPTY DEFAULT:} If the cases do not explicitly show a new drug class being added to treat the target diagnosis, your \texttt{common\_additions} MUST be empty \texttt{[]}.
        \item \textbf{No Invention:} Rely ONLY on the provided text and copy exact pharmacological class names.
    \end{enumerate}
    
    \textbf{OUTPUT FORMAT:} Output ONLY valid JSON. \\
    \texttt{\{"dominant\_pattern": "ADD", "common\_additions": ["Drug Class A"], "reasoning": "Brief 1-sentence reason based on cases."\}}

    \tcblower
    \scriptsize
    
    \textbf{User Prompt:} \\
    \textbf{Current Patient Diagnoses:} \{diagnoses\} \\
================================================== \\
TARGET CLINICAL DIAGNOSIS: $>>>$ \{focus\_txt\} $<<<$ \\
* IMPORTANT: Filter the 'Similar Patient Cases' below. Only pay attention to how physicians treated THIS EXACT TARGET. \\
* Ignore medications given for other unrelated diagnoses in those cases. \\
================================================== \\
\textbf{Similar Patient Cases:} \{rag\_patients\} \\
    
    Analyze explicitly what was ADDED. Output JSON only.
    \end{tcolorbox}
    \vspace{-0.4cm}
    \caption{\textbf{Full prompts for Prescribing Tendency Analysis (Stage 2).} 
    By enforcing a strict \textit{Causality Check}, this stage deliberately filters out high-frequency chronic medications. Furthermore, the \textit{Cautious Empty Default} prevents hallucination by requiring concrete evidence of a new intervention to populate the \texttt{common\_additions} list, effectively neutralizing majority bias.}
    \label{fig:full_prompt_analyzer}
\end{figure*}

\begin{figure*}[p]
    \centering
    \begin{tcolorbox}[colback=parkinson!8!white, colframe=parkinson!80!black, title={
    Evidence-Constrained Prescription Refinement Prompt:
    Parkinson's Disease }, fonttitle=\bfseries]
    \scriptsize
    \textbf{System Prompt:} \\
    You are a strict Clinical Auditor checking a Draft Prescription against RAG Evidence. \\
    You MUST execute your task mechanically using this exact 2-step algorithm. Do not overthink. \\
    ===================================================================== \\
    \textbf{\#\#\# DRUG FILTERING ALGORITHM} \\
    \textbf{STEP 1: PRESERVE ACTIVE HISTORY (MANDATORY)} \\
    Look at the \texttt{Active History} list provided in the prompt.
    \begin{itemize}[leftmargin=*, noitemsep, topsep=0pt]
        \item If \texttt{Active History} is empty or ``None'', you MUST NOT use the ``KEPT'' action. Skip to Step 2.
        \item You MUST put EVERY single drug from \texttt{Active History} into your \texttt{final\_prescription} array.
        \item You MUST create an \texttt{\{"action": "KEPT", "drug": "..."\}} entry in the \texttt{audit\_log} for EACH of these drugs.
        \item NEVER miss or drop a drug from the Active History.
    \end{itemize}

    \textbf{STEP 2: EVALUATE DRAFT DRUGS} \\
    Look at the \texttt{Initial Draft Prescription}. For each drug that is NOT already in Active History:
    \begin{itemize}[leftmargin=*, noitemsep, topsep=0pt]
        \item Is it exactly listed in the RAG \texttt{common\_additions} array?
        \item $\rightarrow$ YES: ADD it. (action: ``ADDED'')
        \item $\rightarrow$ NO: REMOVE it. (action: ``REMOVED'')
    \end{itemize}
    ===================================================================== \\
    \textbf{\#\#\# CRITICAL RULES}
    \begin{itemize}[leftmargin=*, noitemsep, topsep=0pt]
        \item \textbf{STRICT ADD GATE:} You can ONLY add a new drug if it is explicitly listed in \texttt{common\_additions}.
        \item \textbf{SYNC RULE:} If you log ``REMOVED'', it MUST NOT be in \texttt{final\_prescription}.
        \item \textbf{EMPTY FALLBACK:} If \texttt{final\_prescription} is completely empty, pick drug from RAG's \texttt{common\_additions} and ADD it.
    \end{itemize}

    \textbf{\#\#\# REQUIRED JSON FORMAT (DO NOT ADD OTHER KEYS)} \\
    \texttt{\{"final\_prescription": ["Drug A", "Drug B"], "audit\_log": [...], "final\_description": "..."\}}
    
    \tcblower
    \scriptsize
    
    \textbf{User Prompt:} \\
    \textbf{Patient Input:} Subjective: \{subjective\}, Objective: \{objective\}, Assessment: \{assessment\} \\
    \textbf{Active History (Currently taking. ALWAYS KEEP THESE):} \{active\_history\} \\
    \textbf{Initial Draft Prescription (WARNING: This is just a guess. Default action is REMOVE unless proven by RAG):} \{initial\_prescription\} \\
    \{primary\_focus\_block\} \\
    \textbf{RAG Focus Tendencies (what the DB returned for those/similar symptoms; ordered by priority):} \{rag\_focus\_tendency\} \\
    \textbf{Past Clinical Visits:} \{recent\_visit\_history\_text\} \\
    \textbf{FINAL EXECUTION TASK:} Apply the 2-STEP DRUG FILTERING ALGORITHM. 1. Force-copy ALL Active History classes to `final\_prescription` and log them as "KEPT" (unless history is empty).
2. Evaluate remaining Draft classes strictly against RAG `common\_additions`, adopting the specific RAG terminology.
    \end{tcolorbox}

    \vspace{-4mm}

    \begin{tcolorbox}[colback=mimic!8!white, colframe=mimic!90!black, title={
    Evidence-Constrained Prescription Refinement prompt:
    MIMIC-IV }, fonttitle=\bfseries]
    \scriptsize
    \textbf{System Prompt:} \\
    You are a strict Clinical Auditor checking a Draft Prescription against RAG Evidence for complex hospital inpatients (MIMIC-IV). \\
    You MUST execute your task mechanically using this exact 2-step algorithm. Do not overthink. \\
    ===================================================================== \\
    \textbf{\#\#\# DRUG CLASS FILTERING ALGORITHM} \\
    \textbf{STEP 1: PRESERVE ACTIVE HISTORY (MANDATORY)} \\
    Look at the \texttt{Active History} list provided in the prompt.
    \begin{itemize}[leftmargin=*, noitemsep, topsep=0pt]
        \item If \texttt{Active History} is empty or ``None'', you MUST NOT use the ``KEPT'' action. Skip to Step 2.
        \item Otherwise, you MUST put EVERY single drug class from \texttt{Active History} into your \texttt{final\_prescription} array.
        \item You MUST create an \texttt{\{"action": "KEPT", "drug": "..."\}} entry in the \texttt{audit\_log} for EACH of these classes.
        \item NEVER miss or drop a class from the Active History.
    \end{itemize}
    \textbf{STEP 2: EVALUATE DRAFT DRUG CLASSES} \\
    Look at the \texttt{Initial Draft Prescription}. For each drug class that is NOT already in Active History:
    \begin{itemize}[leftmargin=*, noitemsep, topsep=0pt]
        \item Is this exact class name present in ANY focus's RAG \texttt{common\_additions} array?
        \item $\rightarrow$ YES: ADD it to \texttt{final\_prescription}. (action: ``ADDED'')
        \item $\rightarrow$ NO: REMOVE it from the Draft. (action: ``REMOVED'')
    \end{itemize}
    For EVERY Draft drug class you process in Step 2, you MUST create exactly one \texttt{audit\_log} entry (``ADDED'' or ``REMOVED''). If the Initial Draft is non-empty, \texttt{audit\_log} MUST NOT be empty. \\
    ===================================================================== \\
    \textbf{\#\#\# CRITICAL RULES}
    \begin{itemize}[leftmargin=*, noitemsep, topsep=0pt]
        \item \textbf{STRICT ONTOLOGY:} NEVER output broad/umbrella categories (e.g., ``Antihypertensives'', ``Analgesics'', ``Anticoagulants''). ALWAYS keep the exact ATC class names that appear in Active History or RAG (e.g., ``Heparin group'', ``Anilides'').
        \item \textbf{STRICT ADD GATE:} You can ONLY add a new drug class if it is \textbf{literally present} (exact string match) in RAG \texttt{common\_additions} for at least one focus.
        \item \textbf{SYNC RULE:} If you log ``REMOVED'' for a class, that class MUST NOT appear anywhere in \texttt{final\_prescription}. If a class remains in \texttt{final\_prescription}, it MUST have action ``KEPT'' or ``ADDED'' in \texttt{audit\_log}.
        \item \textbf{COVERAGE RULE:} When the Initial Draft is non-empty, every Draft class that is not in Active History MUST appear exactly once in \texttt{audit\_log} as either ``ADDED'' or ``REMOVED''.
        \item \textbf{EMPTY FALLBACK:} If Active History is empty AND you removed all Draft classes AND at least one RAG \texttt{common\_additions} class exists, you may ADD some clearly relevant class from RAG.
    \end{itemize}

    \textbf{\#\#\# REQUIRED JSON FORMAT (DO NOT ADD OTHER KEYS)} \\
    \texttt{\{"final\_prescription": ["Drug A", "Drug B"], "audit\_log": [...], "final\_description": "..."\}}

    \tcblower
    \scriptsize
    
    \textbf{User Prompt:} \\
    \textbf{Patient Input (Diagnoses):} \{diagnoses\} \\
    \textbf{Active History (Currently taking. ALWAYS KEEP THESE):} \{active\_history\} \\
    \textbf{Initial Draft Prescription (WARNING: This is just a guess. Default action is REMOVE unless proven by RAG):} \{initial\_prescription\} \\
    \textbf{RAG Focus Tendencies (what the DB returned for those/similar symptoms; ordered by priority):} \{rag\_focus\_tendency\} \\
    \textbf{Past Clinical Visits:} \{recent\_visit\_history\_text\} \\
    \textbf{FINAL EXECUTION TASK:} Apply the 2-STEP DRUG CLASS FILTERING ALGORITHM.
1. Force-copy ALL Active History classes to `final\_prescription` and log them as "KEPT" (unless history is empty).
2. Evaluate remaining Draft classes strictly against RAG `common\_additions`, adopting the specific RAG terminology.
    \end{tcolorbox}
    \vspace{-5mm}
    \caption{\textbf{Full prompts for 
    Evidence-Constrained Prescription Refinement 
    (Stage 3).} 
    Step 1 prioritizes treatment stability by preserving existing history,
while Step 2 adds new medications only when they are supported by similar patient cases.
    }
    \label{fig:full_prompt_verifier}
\end{figure*}


\begin{figure*}[p]
    \centering
    \begin{tcolorbox}[colback=violet!5!white, colframe=violet!75!black, title={Clinical Summary Generator Prompt }, fonttitle=\bfseries]
    \scriptsize
    \textbf{System Prompt:} \\
    You are a Senior Clinical Consultant summarizing cases for attending physicians. \\
    Format your output EXACTLY as follows: \\
    
    \textbf{* Patient summary *} \\
    (Detailed description of patient's current symptoms, stability, and clinical trajectory in 2-3 sentences.) \\
    
    \textbf{* Key word *} \\
    (Comma-separated list of clinical focus keywords extracted from symptoms/diagnoses.) \\
    
    \textbf{* Clinical Evidence *} \\
    (Concise summary of treatment patterns found in similar historical cases. Mention standard approaches like maintaining or adjusting medications and specific evidence found for the current clinical focus.) \\
    
    \textbf{* Prescribe *} \\
    (Drug Name) : (Detailed clinical rationale. Explain if the drug is being continued from the patient's history, why it was added or adjusted based on current symptoms, and how patterns from similar cases or clinical validation influenced the final recommendation.) \\
    
    \textbf{Rules:}
    \begin{itemize}[leftmargin=*, noitemsep, topsep=0pt]
        \item Use ONLY provided inputs. No hallucinations.
        \item Ensure the * Prescribe * section covers EVERY drug in the final recommendation list.
        \item No internal monologue or \texttt{<think>} tags.
        \item Output plain text structure only.
    \end{itemize}
    
    \textbf{User Prompt:} \\
    **Patient State:** \{patient\_state\} \\
    **Initial Draft (Proposed Plan):** \{initial\_prescription\} \\
    **Clinical Evidence from Similar Cases:** \{rag\_tendency\_by\_focus\} \\
    **Clinical Validation Log:** \{audit\_log\} \\
    **Final Recommended Medications:** \{final\_answer\_list\}
    \end{tcolorbox}
    \vspace{-4mm}
    \caption{\textbf{Full prompt for Clinical Summary Generation (Stage 4).} 
    This stage integrates diverse evidence—including patient history, prescribing tendency analysis, and refined tendencies—into a cohesive clinical narrative. 
    As a result, this stage provides clinicians with both the final drug recommendations and the clinical reasoning behind them.
    }
    \label{fig:full_prompt_summary}
\end{figure*}

\begin{figure*}[htbp]
    \centering
    \begin{tcolorbox}[colback=parkinson!8!white, colframe=parkinson!80!black, title={Evaluator Prompt for Keyword Extractor}, fonttitle=\bfseries]
    \scriptsize
    \textbf{System Prompt:} \\
    You are a clinical evaluator for a Parkinson's disease prescription support system. \\
    
    \textbf{What the keyword extractor does (the system you are evaluating):}
    \begin{itemize}[leftmargin=*, noitemsep, topsep=0pt]
        \item It is a Medical Symptom Extractor. It identifies \textbf{NEW, WORSENING, or UNRESOLVED} pathological symptoms that \textbf{require a medication change}.
        \item It must output \textbf{empty keywords} when: the patient is STABLE, IMPROVING, or ``feeling better''; or when the text is just conversational (e.g. ``test results explained'', ``stopped drinking'') with no acute symptom. So ``no keywords'' is correct when the case does not need RAG (no acute symptom to focus on).
        \item When it does extract, it uses \textbf{literal phrases from the note}, at most 2 phrases, focused on \textbf{acute/severe symptoms} --- not broad diagnoses alone (e.g. not ``Parkinson's disease'' by itself).
    \end{itemize}
    
    \vspace{2mm}
    \textbf{You will be given:}
    \begin{enumerate}[leftmargin=*, noitemsep, topsep=0pt]
        \item Current clinical note (\texttt{\{patient\_input\}}) and optional previous visit history (\texttt{\{history\_soap\}}, max 3 visits).
        \item Ground-truth prescription list (ingredient names) for this visit.
        \item ``Focus areas'': the keywords our extractor output from Subjective/Assessment, \textbf{or} a notice that it output no keywords.
    \end{enumerate}
    
    \vspace{2mm}
    \textbf{Your task (two cases):}
    \begin{itemize}[leftmargin=*, noitemsep, topsep=0pt]
        \item \textbf{When keywords were extracted:} Evaluate how well the prescription aligns with those focus keywords (do the drugs address the extracted symptoms?). Score 1-5.
        \item \textbf{When no keywords were extracted:} Judge whether that was correct under the extractor's rules --- i.e. was this case actually stable/improving/conversational with no acute symptom (then 5 = correct not to extract), or was there a new/worsening/unresolved symptom that should have been extracted (then 1-2 = wrong not to extract)? Score 1-5.
    \end{itemize}
    
    \vspace{2mm}
    \textbf{Output in this format (exact headers):} \\
    \textbf{Relevance (1-5):} [Integer 1-5] \\
    \textbf{Explanation:} [2-4 sentences] \\
    \textbf{Summary:} [One sentence] \\
    Do not prescribe or suggest drugs; only evaluate. \\

    \tcblower
    \scriptsize
    
    \textbf{User Prompt:} \\
    \textbf{Current Clinical Note (patient\_input):} \\
    \{patient\_input\} \\
    
    \textbf{Previous Visit History (SOAP):} \\
    \{history\_soap\} \\
    
    \textbf{Ground-Truth Prescription (ingredient list):} \\
    \{gt\_txt\} \\
    
    \textbf{Focus areas (keywords our system extracted, or notice that none were extracted):} \\
    \{focus\_desc\} \\
    
    \vspace{1mm}
    \textit{The extractor outputs keywords only for NEW/WORSENING/UNRESOLVED symptoms that need a medication change; it must output none when the case is stable/improving or conversational with no acute symptom. For this case: if keywords were extracted, evaluate prescription vs those keywords; if none were extracted, judge whether that was correct (stable/no acute = correct, clear acute symptom missed = wrong). Output \textbf{Relevance (1-5)}, \textbf{Explanation}, and \textbf{Summary} as specified.}
    \end{tcolorbox}
    \vspace{-4mm}
    \caption{\textbf{Prompt template for the automated clinical evaluator (LLM-as-a-judge) used to assess the Focus Query Extractor's performance.} The evaluator is instructed to verify not only the clinical relevance of the extracted keywords against the ground-truth prescriptions, but more importantly, to strictly penalize the model if it fails to apply the `Empty Rule' for stable or conversational cases.}
    \label{fig:prompt_keyword_evaluator}
\end{figure*}

\begin{figure*}[p]
    \centering
    \begin{tcolorbox}[colback=gray!5!white, colframe=gray!75!black, title={Baseline Similar Patients RAG (TreatRAG): Parkinson's Disease}, fonttitle=\bfseries]
    \scriptsize
    \textbf{System Prompt:} \\
    You are a Parkinson's disease medication specialist. \\
    
    Make your prescription based on the patient's current symptoms and current medications.
    If there is active history, start from that active history! 
    Then, maintain(if there is no new symptoms or reason to modify drugs) or add or remove drugs based on the patient's current symptoms and current medications.
    You should be careful - there should be certain reason for adding or removing drugs based on the patient's current symptoms and current medications. \\
    
    When similar cases are provided, use them only as a secondary reference.
    Always prioritize SUBJECTIVE/OBJECTIVE/ASSESSMENT and active history. \\
    
    You should put one drug name at every line! \\
    \textbf{FORMAT STRICT:} Output ONLY the [START]...[END] block. No extra text. \\
    
    \textbf{OUTPUT:} \\
    \texttt{[START]} \\
    \texttt{(Drug Name) | (short reason in 10 words or less)} \\
    \texttt{(Drug Name) | (short reason in 10 words or less)} \\
    \texttt{[END]} \\

    \tcblower
    \scriptsize
    
    \textbf{User Prompt:} \\
    \textbf{Clinical Note:} \\
    - Subjective: \{subjective\} \\
    - Objective: \{objective\} \\
    - Assessment: \{assessment\} \\
    
    \textbf{History Summary:} \\
    \{history\} \\
    
    \textbf{Past Clinical Visits (up to 3 visits before current):} \\
    \{recent\_visit\_history\_text\} \\
    
    \textbf{Similar Cases (top retrieved):} \\
    \{similar\_cases\} \\
    
    \textbf{Task:} Generate prescription. Should put at least one drug in output.
    \end{tcolorbox}

    \vspace{-0.4cm}

    \begin{tcolorbox}[colback=gray!5!white, colframe=gray!75!black, title={Baseline Similar Patients RAG (TreatRAG): MIMIC-IV}, fonttitle=\bfseries]
    \scriptsize
    \textbf{System Prompt:} \\
    You are a clinical medication specialist for complex hospital inpatients (MIMIC-IV dataset). \\
    
    Make your prescription based on the patient's current diagnoses and most recent medications.
    If there is a most recent medications list, start from that list.
    Then, maintain(if there is no new clinical reason to modify drugs) or add or remove drug classes based on the patient's diagnoses and overall clinical status.
    You should be careful - there should be certain reason for adding or removing drug classes based on the patient's diagnoses and medication history. \\
    
    When similar cases are provided, use them only as a secondary reference.
    Always prioritize CURRENT DIAGNOSES and Most Recent Medications. \\
    
    Use pharmacological/therapeutic class names (not specific drug brands or single ingredients), and copy class names exactly as they appear in the input (do not invert words or invent umbrella categories). \\
    
    You should put one drug class name at every line! \\
    \textbf{FORMAT STRICT:} Output ONLY the [START]...[END] block. No extra text. \\
    
    \textbf{STABILITY RULE (no most recent medications list):}
    \begin{itemize}[leftmargin=*, noitemsep, topsep=0pt]
        \item If the clinical picture is stable/improved/no major new problems, output ONLY a conservative minimal set of classes.
        \item Prefer continuing existing chronic medication classes when they are clearly indicated.
        \item Avoid starting broad new classes unless there is clear diagnostic support.
    \end{itemize}
    
    \textbf{OUTPUT:} \\
    \texttt{[START]} \\
    \texttt{(Drug Class Name) | (short reason in 10 words or less)} \\
    \texttt{(Drug Class Name) | (short reason in 10 words or less)} \\
    \texttt{[END]} \\

    \tcblower
    \scriptsize
    
    \textbf{User Prompt:} \\
    \textbf{Patient Information:} \\
    - Diagnoses: \{diagnoses\} \\
    - Most Recent Medications: \{medications\} \\
    
    \textbf{Past Clinical Visits (up to 3 visits before current):} \\
    \{recent\_visit\_history\_text\} \\
    
    \textbf{Similar Cases (top retrieved):} \\
    \{similar\_cases\} \\
    
    \textbf{Task:} Generate prescription. Should put at least one drug class in output.
    \end{tcolorbox}
    \vspace{-0.4cm}
    \caption{\textbf{Prompt designs for the Baseline Similar Patients RAG (TreatRAG).} 
    In this method, retrieved similar historical cases are directly appended to the prompt context. 
    }
    \label{fig:prompt_treatrag_baseline}
\end{figure*}

\begin{figure*}[p]
    \centering
    \begin{tcolorbox}[colback=gray!5!white, colframe=gray!75!black, title={Baseline Guideline RAG: Parkinson's Disease}, fonttitle=\bfseries]
    \scriptsize
    \textbf{System Prompt:} \\
    You are a Parkinson's disease medication specialist. \\
    
    Make your prescription based on the patient's current symptoms and current medications.
    If there is active history, start from that active history!
    Then, maintain(if there is no new symptoms or reason to modify drugs) or add or remove drugs based on the patient's current symptoms.
    You should be careful - there should be certain reason for adding or removing drugs based on the patient's current symptoms and current medications. \\
    
    \textbf{GUIDELINE RULE:}
    \begin{itemize}[leftmargin=*, noitemsep, topsep=0pt]
        \item If `Similar Guidelines' are provided, apply them if they match the patient's specific symptoms.
        \item If none are relevant or empty, completely ignore them and rely on your clinical judgment.
    \end{itemize}
    
    \textbf{STABILITY RULE (no active history):}
    \begin{itemize}[leftmargin=*, noitemsep, topsep=0pt]
        \item If symptoms are stable/improved/no major worsening, output ONLY one conservative drug.
        \item Prefer a non-escalating option from guidelines when available.
        \item Avoid starting Levodopa/Carbidopa unless there is clear worsening/functional decline.
    \end{itemize}
    
    You should put one drug name at every line! \\
    \textbf{FORMAT STRICT:} Output ONLY the [START]...[END] block. No extra text. \\
    
    \textbf{OUTPUT:} \\
    \texttt{[START]} \\
    \texttt{(Drug Name) | (short reason in 10 words or less)} \\
    \texttt{(Drug Name) | (short reason in 10 words or less)} \\
    \texttt{[END]}

    \tcblower
    \scriptsize
    
    \textbf{User Prompt:} \\
    \textbf{Clinical Note:} \\
    - Subjective: \{subjective\} \\
    - Objective: \{objective\} \\
    - Assessment: \{assessment\} \\
    
    \textbf{Most Recent Medications:} \{history\} \\
    
    \textbf{Past Clinical Visits (up to 3 visits before current):} \\
    \{recent\_visit\_history\_text\} \\
    
    \textbf{Similar Guidelines:} \\
    \{similar\_guidelines\} \\
    
    \textbf{Task:} Generate prescription. Should put at least one drug in output.
    \end{tcolorbox}

    \vspace{-0.4cm}

    \begin{tcolorbox}[colback=gray!5!white, colframe=gray!75!black, title={Baseline Guideline RAG: MIMIC-IV}, fonttitle=\bfseries]
    \scriptsize
    \textbf{System Prompt:} \\
    You are a clinical medication specialist for complex hospital inpatients (MIMIC-IV dataset). \\
    
    Make your prescription based on the patient's current diagnoses and most recent medications.
    If there is a most recent medications list, start from that list.
    Then, maintain(if there is no new clinical reason to modify drugs) or add or remove drug classes based on the patient's diagnoses and overall clinical status.
    You should be careful - there should be certain reason for adding or removing drug classes based on the patient's diagnoses and medication history. \\
    
    \textbf{GUIDELINE RULE:}
    \begin{itemize}[leftmargin=*, noitemsep, topsep=0pt]
        \item If `Similar Guidelines' are provided, apply them if they match the patient's specific diagnoses.
        \item If none are relevant or empty, completely ignore them and rely on your clinical judgment.
    \end{itemize}
    
    Use pharmacological/therapeutic class names (not specific brands/ingredients), and copy class names literally when possible. \\
    
    You should put one drug class name at every line! \\
    \textbf{FORMAT STRICT:} Output ONLY the [START]...[END] block. No extra text. \\
    
    \textbf{OUTPUT:} \\
    \texttt{[START]} \\
    \texttt{(Drug Class Name) | (short reason in 10 words or less)} \\
    \texttt{(Drug Class Name) | (short reason in 10 words or less)} \\
    \texttt{[END]}

    \tcblower
    \scriptsize
    
    \textbf{User Prompt:} \\
    \textbf{Patient Information:} \\
    - Diagnoses: \{diagnoses\} \\
    - Most Recent Medications: \{medications\} \\
    
    \textbf{Past Clinical Visits (up to 3 visits before current):} \\
    \{recent\_visit\_history\_text\} \\
    
    \textbf{Similar Guidelines:} \\
    \{similar\_guidelines\} \\
    
    \textbf{Task:} Generate prescription. Should put at least one drug class in output.
    \end{tcolorbox}
    \vspace{-0.4cm}
    \caption{\textbf{Prompt designs for the Baseline Guideline RAG.}
    This represents a traditional retrieval-augmented generation setup where retrieved excerpts from general clinical guidelines or literature are injected directly into the prompt context. 
    }
    \label{fig:prompt_guideline_baseline}
\end{figure*}

\begin{figure*}[p]
    \centering
    \begin{tcolorbox}[colback=gray!5!white, colframe=gray!75!black, title={MedReflect Stage 1: Question Generation}, fonttitle=\bfseries]
    \scriptsize
    \textbf{System Prompt:} \\
    You are a professional doctor.

    \tcblower
    \scriptsize
    
    \textbf{User Prompt:} \\
    Here is a medical query from your patient: \\
    \texttt{<Query>:} \\
    \texttt{[Subjective]:} \{subjective\} \\
    \texttt{[Objective]:} \{objective\} \\
    \texttt{[Assessment]:} \{assessment\} \\
    \texttt{[Active History]:} \{active\_history\} \\
    \texttt{</Query>} \\
    
    Here is the patient's previous prescription (max 3): \\
    \{previous\_prescription\} \\
    
    Here is your response for the medical query: \\
    \texttt{<Response>:}\{initial\_prescription\}\texttt{</Response>} \\
    
    The initial answer in your response might be incorrect or incomplete (e.g., missing active history drugs or adding unsupported drugs), so you need to ask a reflective question based on your query and your response. \\
    
    Now please provide a brief question(Strictly follow this format:\texttt{<Reflective Question>}your response\texttt{</Reflective Question>}):
    \end{tcolorbox}

    \vspace{-0.4cm}

    \begin{tcolorbox}[colback=gray!5!white, colframe=gray!75!black, title={MedReflect Stage 2: Answer Generation}, fonttitle=\bfseries]
    \scriptsize
    \textbf{System Prompt:} \\
    You are a professional doctor.

    \tcblower
    \scriptsize
    
    \textbf{User Prompt:} \\
    Here is a medical query from your patient: \\
    \texttt{<Query>:} \\
    \texttt{[Subjective]:} \{subjective\} \\
    \texttt{[Objective]:} \{objective\} \\
    \texttt{[Assessment]:} \{assessment\} \\
    \texttt{[Active History]:} \{active\_history\} \\
    \texttt{</Query>} \\
    
    Here is the patient's previous prescription (max 3): \\
    \{previous\_prescription\} \\
    
    Here is your response for the medical query: \\
    \texttt{<Response>:}\{initial\_prescription\}\texttt{</Response>} \\
    
    Here is your own reflective question for your response: \\
    \texttt{<Reflective Question>:}\{reflective\_question\}\texttt{</Reflective Question>} \\
    
    Now, please provide a concise answer for the reflective question(Strictly follow this format:\texttt{<Reflective Answer>}your response\texttt{</Reflective Answer>}):
    \end{tcolorbox}

    \vspace{-0.4cm}

    \begin{tcolorbox}[colback=gray!5!white, colframe=gray!75!black, title={MedReflect Stage 3: Refinement}, fonttitle=\bfseries]
    \scriptsize
    \textbf{System Prompt:} \\
    You are a professional doctor.

    \tcblower
    \scriptsize
    
    \textbf{User Prompt:} \\
    Here is a medical query from your patient: \\
    \texttt{<Query>:} \\
    \texttt{[Subjective]:} \{subjective\} \\
    \texttt{[Objective]:} \{objective\} \\
    \texttt{[Assessment]:} \{assessment\} \\
    \texttt{[Active History]:} \{active\_history\} \\
    \texttt{</Query>} \\
    
    Here is the patient's previous prescription (max 3): \\
    \{previous\_prescription\} \\
    
    Here is your own reflection on your initial answer: \\
    \texttt{<Self-Reflection>:}\{reflective\_question\} \{reflective\_answer\}\texttt{</Self-Reflection>} \\
    
    Here is the response you need to complete(Complete each blank): \\
    \texttt{<Response>:}The final prescribed medications are: \texttt{<mask, type: drug\_list></Response>} \\
    
    Now, according on this reflection, your completed answer is(Strictly follow this format:\texttt{<Answer>:}your refine entity,eg:[Drug A, Drug B]\texttt{</Answer>}):
    \end{tcolorbox}
    \vspace{-0.4cm}
    \caption{\textbf{Prompt designs for the MedReflect baseline.} The self-reflection process is divided into three sequential stages: 
    (1) \textbf{Question Generation}, where the model critiques its initial answer; 
    (2) \textbf{Answer Generation}, where it resolves the critique; 
    and (3) \textbf{Refinement}, where it outputs the final corrected prescription based on its internal reflection.}
    \label{fig:prompt_medreflect}
\end{figure*}

\begin{figure*}[p]
    \centering
    
    \begin{tcolorbox}[colback=mimic!8!white, colframe=mimic!90!black, title={(Simplified Stage 1) Focus Query Extractor: MIMIC-IV}, fonttitle=\bfseries]
    \scriptsize
    \textbf{System Prompt:} \\
    Extract the most clinically relevant diagnosis phrases from a MIMIC-IV patient's diagnosis list.
    Return up to 2 phrases that are active and likely to drive medication decisions.
    Use exact wording from the input. If none are relevant, return an empty list. \\
    
    \textbf{Output JSON only:} \\
    \texttt{\{"keywords": ["phrase1", "phrase2"]\}}

    \tcblower
    \scriptsize

    \textbf{User Prompt:} \\
    \textbf{Diagnoses:} \\
    \{text\} \\
    \textbf{Current medications:} \\
    \{active\_history\} \\
    Return JSON only.
    \end{tcolorbox}
    \vspace{-0.4cm}

    \begin{tcolorbox}[colback=mimic!8!white, colframe=mimic!90!black, title={(Simplified Stage 2) Prescribing Tendency Analysis: MIMIC-IV}, fonttitle=\bfseries]
    \scriptsize
    \textbf{System Prompt:} \\
    You are a clinical pattern analyzer for MIMIC-IV inpatients. \\
    
    From the provided similar patient cases, identify which medication classes were newly added for the given diagnosis. Use exact class names from the cases. If evidence is weak or unclear, return an empty list. \\

    \textbf{Output JSON only:} \\
    \texttt{\{"dominant\_pattern": "ADD", "common\_additions": ["Drug Class A"], "reasoning": "One sentence."\}}

    \tcblower
    \scriptsize

    \textbf{User Prompt:} \\
    \textbf{Patient diagnoses:} \{diagnoses\} \\
    \{keyword\_focus\_instruction\} \\
    \textbf{Similar cases:} \{rag\_patients\} \\
    Output JSON only.
    \end{tcolorbox}
    \vspace{-0.4cm}

    \begin{tcolorbox}[colback=mimic!8!white, colframe=mimic!90!black, title={(Simplified Stage 3) Evidence-Constrained Prescription Refinement prompt: MIMIC-IV}, fonttitle=\bfseries]
    \scriptsize
    \textbf{System Prompt:} \\
    You are a clinical pharmacist reviewing a draft prescription for a MIMIC-IV inpatient. \\
    Build the final prescription by applying these steps in order: \\

    \textbf{STEP 1 — PRESERVE ACTIVE HISTORY} \\
    Place every class from \texttt{Active History} into \texttt{final\_prescription}. \\
    Log each as \texttt{\{"action": "KEPT", "drug": "...", "reason": "In active history."\}}. \\
    Include ALL Active History items, even those not in the Draft. \\

    \textbf{STEP 2 — EVALUATE THE DRAFT} \\
    For each class in the Draft that is not already in \texttt{final\_prescription} (from Step 1):
    \begin{itemize}[leftmargin=*, noitemsep, topsep=0pt]
        \item If it appears in \texttt{RAG Focus Tendencies} (common\_additions for any diagnosis): include it, log as "ADDED".
        \item Otherwise: exclude it, log as "REMOVED".
    \end{itemize}
    Use exact class names from Active History or RAG. Do not rename or generalize. \\

    \textbf{Output JSON only:} \\
    \texttt{\{ \\
    ~ "final\_prescription": ["Drug Class A", "Drug Class B"], \\
    ~ "audit\_log": [ \\
    ~ ~ \{"action": "KEPT", "drug": "Drug Class A", "reason": "In active history."\}, \\
    ~ ~ \{"action": "ADDED", "drug": "Drug Class B", "reason": "In RAG common\_additions for [focus]."\}, \\
    ~ ~ \{"action": "REMOVED", "drug": "Drug Class C", "reason": "Not in RAG evidence."\} \\
    ~ ], \\
    ~ "final\_description": "One sentence summary." \\
    \}}

    \tcblower
    \scriptsize

    \textbf{User Prompt:} \\
    \textbf{Diagnoses:} \{diagnoses\} \\
    \textbf{Active History (include ALL of these in final\_prescription):} \\
    \{active\_history\} \\
    \textbf{Draft Prescription:} \\
    \{initial\_prescription\} \\
    \textbf{RAG Focus Tendencies:} \\
    \{rag\_focus\_tendency\} \\
    \textbf{Past Visits:} \\
    \{recent\_visit\_history\_text\} \\
    Output JSON only.
    \end{tcolorbox}
    \vspace{-0.4cm}
    
    \caption{\textbf{Simplified prompts used in the robustness experiment (Appendix~\ref{app:robustness_prompt}).} }
    \label{fig:simplified_prompts}
\end{figure*}


\begin{figure*}[p]
    \centering
    \begin{tcolorbox}[colback=gray!5!white, colframe=black!70!white, title={Clinical Relevance Judge for Stage~2 Medication Additions}, fonttitle=\bfseries]
    \scriptsize
    \textbf{System Prompt:} \\
    You are a clinical pharmacology expert reviewing medication additions extracted from a RAG-based prescribing tendency analysis. The ``focus'' is a symptom or diagnosis keyword used to retrieve similar patients. The ``medication'' is a drug class that was commonly ADDED in those similar cases. Patient context (when provided) may list co-occurring conditions --- use it. \\
    \\
    Important: focus is a retrieval key, not a claim that the drug must specifically treat that exact phrase. Similar patients are often complex and receive concurrent disease therapy, supportive care, prophylaxis, symptom control, and comorbidity medications. Prefer clinically inclusive judgments without marking obvious noise as relevant. \\
    \\
    Mark ``relevant'': TRUE if ANY of the following apply: \\
    (a) The medication directly treats or manages the focus symptom/diagnosis, OR a common complication of it. \\
    (b) The medication is standard disease-modifying or symptomatic therapy for the disease implied by the focus or patient context (e.g., antiparkinsonian agents when the focus is a Parkinson motor/non-motor symptom, wearing-off, tremor, gait, constipation in a PD note, vertigo/dizziness in a PD note). Do NOT require the drug to treat that exact symptom word-for-word. \\
    (c) Routine supportive / peri-procedural / hospital-course care commonly used for patients with this focus (e.g., DVT prophylaxis/heparin, antiemetics/5HT3 antagonists, PPIs, analgesics/anilides, laxatives, electrolytes, topical/oral care agents, cardiac supportive agents). \\
    (d) The medication treats a common comorbidity or co-occurring problem associated with this focus, including conditions listed in patient context. \\
    (e) External-cause / activity / accident / fall-type focuses: treat as the reason for the encounter --- mark TRUE for typical trauma/ortho/neuro care and supportive meds. \\
    (f) Antibiotics / local anti-infectives when infection, surgery, or mucosal injury is plausible for this focus or co-listed diagnoses. \\
    (g) The medication is prescribed reactively to a side effect of other treatments for this condition. \\
    (h) Patient context lists another diagnosis that this medication standardly treats. \\
    \\
    Mark ``relevant'': FALSE ONLY if: \\
    -- There is no plausible clinical connection to the focus, implied disease, OR co-listed conditions; OR \\
    -- The label is nonsensical/non-drug; OR \\
    -- The drug is highly specialty-specific with zero overlap to the clinical picture. \\
    Do NOT mark FALSE merely because the link is indirect, prophylactic, symptomatic, comorbidity-driven, or because the drug treats the broader disease rather than the exact focus wording. \\
    \\
    Default to TRUE when uncertain. Only mark FALSE when you are confident there is no connection. \\
    You MUST return exactly one judgment per pair, in the same order as the input. \\
    \\
    Output format (REQUIRED --- return a JSON object with key ``judgments''): \\
    \texttt{\{"judgments": [\{"relevant": true/false, "confidence": "high"/"medium"/"low", "reason": "<one short sentence>"\}, ...]\}}

    \tcblower
    \scriptsize
    \textbf{User Prompt:} \\
    Dataset: \{Parkinson disease $\mid$ MIMIC-IV inpatient\} \\
    Focus is a retrieval keyword only; medications may reflect disease therapy, concurrent care, or comorbidities. Mark TRUE for standard disease therapy even if it does not treat the exact focus wording; mark FALSE only when there is clearly no clinical connection. \\
    \\
    Judge each pair below: \\
    \\
    1. Focus: \{focus\} \\
    \quad Medication: \{medication\} \\
    \quad Patient context: \{patient\_context\} \\
    2. Focus: \{focus\} \\
    \quad Medication: \{medication\} \\
    \quad Patient context: \{patient\_context\} \\
    \ldots \\
    \\
    \textit{Note: up to 5 pairs per API call. Patient context is the first 400 characters of \texttt{patient\_input} (HPH) or \texttt{assessment}/\texttt{patient\_input} (MIMIC-IV), and is omitted if empty.}
    \end{tcolorbox}
    \caption{Prompt for the GPT-4o clinical relevance judge used to evaluate Stage~2 medication additions.}
    \label{fig:prompt_medication_relevance_judge}

\end{figure*}

\end{document}